\documentclass[lettersize,journal]{IEEEtran}
\usepackage{amsmath,amsfonts}
\usepackage{algorithmic}
\usepackage{algorithm}
\usepackage{array}
\usepackage[caption=false,font=footnotesize,labelfont=sf,textfont=sf]{subfig}
\usepackage{textcomp}
\usepackage{stfloats}
\usepackage{url}
\usepackage{hyperref}
\usepackage{verbatim}
\usepackage{graphicx}
\usepackage{cite}
\usepackage{tikz}
\newcommand*\circled[1]{\tikz[baseline=(char.base)]{
            \node[shape=circle,draw,inner sep=1pt] (char) {#1};}}
\usepackage{svg}
\usepackage{todonotes}

\usepackage{setspace}
\usepackage{tablefootnote}
\usepackage{threeparttable}
\usepackage{siunitx}
\usepackage{multirow}
\usepackage[inline]{enumitem}
\newcommand{\vtheta}{\ensuremath{\mathrm{\boldsymbol{\theta}}}}
\newcommand{\vx}{\ensuremath{\mathrm{\bold{x}}}}

\newcommand{\vy}{\ensuremath{\mathrm{\bold{y}}}}
\newcommand{\vb}{\ensuremath{\mathrm{\bold{b}}}}

\newcommand{\mW}{\ensuremath{\mathrm{\bold{W}}}}
\newcommand{\mAlpha}{{\boldsymbol{{\alpha}}}}
\allowdisplaybreaks

\newcommand{\x}{\mathbf{x}}

\newcommand{\y}{\mathbf{y}}
\newcommand{\z}{\mathbf{z}}

\newcommand{\va}{\ensuremath{\mathrm{\bold{a}}}}
\newcommand{\vw}{\ensuremath{\mathrm{\bold{W}}}}

\newcommand{\vnet}{\ensuremath{\mathrm{\bold{f}}}}

\usepackage{titlesec}

\titlespacing*{\section}
{0pt}{1.2ex plus 1ex minus .2ex}{1.ex plus .2ex}
\titlespacing*{\subsection}
{0pt}{1.2ex plus 1ex minus .2ex}{1.ex plus .2ex}

\def\fl[#1\]{\begin{align}#1\end{align}}
\def\[#1\]{\begin{align*}#1\end{align*}}

\def\*[#1\]{\begin{align*}#1\end{align*}}
\begin{document}
\bstctlcite{IEEEexample:BSTcontrol}

\title{\huge Scale-Dropout: Estimating Uncertainty in Deep Neural Networks Using Stochastic Scale}


\author{\IEEEauthorblockN{Soyed Tuhin Ahmed\IEEEauthorrefmark{9}\IEEEauthorrefmark{2}, Kamal Danouchi\IEEEauthorrefmark{3}, Michael Hefenbrock\IEEEauthorrefmark{4}, Guillaume Prenat\IEEEauthorrefmark{3}, Lorena Anghel\IEEEauthorrefmark{3}, Mehdi B. Tahoori\IEEEauthorrefmark{2}\\}
\IEEEauthorblockA{\IEEEauthorrefmark{2}Karlsruhe Institute of Technology, Karlsruhe, Germany, \IEEEauthorrefmark{9}corresponding author, email: soyed.ahmed@kit.edu}
\IEEEauthorblockA{\IEEEauthorrefmark{3}Univ. Grenoble Alpes, CEA, CNRS, Grenoble INP, and IRIG-Spintec, Grenoble, France}
\IEEEauthorblockA{\IEEEauthorrefmark{4}RevoAI GmbH, Karlsruhe, Germany}
\vspace{-2em}
}

\maketitle

\addtolength\abovedisplayskip{-0.em}%
\addtolength\belowdisplayskip{-0.em}%
\setlength{\textfloatsep}{2pt}

\begin{abstract}
Uncertainty estimation in Neural Networks (NNs) is vital in improving reliability and confidence in predictions, particularly in safety-critical applications. Bayesian Neural Networks (BayNNs) with Dropout as an approximation offer a systematic approach to quantifying uncertainty, but they inherently suffer from high hardware overhead in terms of power, memory, and computation. Thus, the applicability of BayNNs to edge devices with limited resources or to high-performance applications is challenging. Some of the inherent costs of BayNNs can be reduced by accelerating them in hardware on a Computation-In-Memory (CIM) architecture with spintronic memories and binarizing their parameters. However, numerous stochastic units are required to implement conventional Dropout-based BayNN. 
In this paper, we propose the Scale Dropout, a novel regularization technique for Binary Neural Networks (BNNs), and Monte Carlo-Scale Dropout (MC-Scale Dropout)-based BayNNs for efficient uncertainty estimation. Our approach requires only one stochastic unit for the entire model, irrespective of the model size, leading to a highly scalable Bayesian NN.
Furthermore, we introduce a novel Spintronic memory-based CIM architecture for the proposed BayNN that achieves more than $100\times$ energy savings compared to the state-of-the-art. We validated our method to show up to $1\%$ improvement in predictive performance and superior uncertainty estimates compared to related works.

\end{abstract}

\addtolength\abovedisplayskip{-0.2em}%
\addtolength\belowdisplayskip{-0.2em}%
\setlength{\textfloatsep}{0pt}
\setlength{\belowcaptionskip}{-3pt}
\captionsetup{belowskip=0pt, font={footnotesize}}
\captionsetup[figure]{font=footnotesize}

\section{Introduction}

In recent years, Neural networks (NNs) have revolutionized various fields of artificial intelligence due to their exceptional ability to learn complex patterns and generate accurate predictions. NN models have produced remarkable results in a variety of tasks, including image recognition~\cite{krizhevsky2012imagenet}, natural language processing~\cite{devlin2018bert}, and are even being deployed in safety-critical applications such as autonomous vehicles, automatic medical diagnosis~\cite{esteva2021deep}, and automated optical inspection in industrial applications~\cite{tang2020anomaly}. Despite the achievements of NNs, they are unable to quantify the uncertainty associated with their predictions. Estimating uncertainty is essential to understand the risks associated with prediction and to make informed decisions, especially in safety-critical domains, where incorrect predictions can have severe consequences~\cite{kendall2017uncertainties}. 

In contrast to traditional NNs, Bayesian Neural Networks (BayNNs) offer a principled approach to uncertainty estimation~\cite{neal2012bayesian}. 
BayNN is a model based on the Bayesian framework for model interpretation and decision-making that introduces probability distributions over the weights or activation of the network. Despite the benefit of BayNN, they inherently have higher hardware overhead in terms of power, memory consumption, throughput, computation, and number of stochastic units representing probability distributions of BayNNs. 
Consequently, the applicability of BayNN is limited in the context of edge devices, for example, microcontrollers and smartphones, where computing and memory resources are limited and power consumption and throughput are a concern~\cite{hubara2016binarized}. 

Furthermore, exact Bayesian inference is often computationally intractable, necessitating the adoption of various approximation techniques. The ensemble methods~\cite{lakshminarayanan2017simple}, and variational inference~\cite{blundell2015weight} are among the prominent methods. However, they have high overhead as a result of the storage of multiple copies of the model, or they have twice the number of parameters compared to traditional NNs. On the other hand, Monte Carlo Dropout (MC-Dropout) based Bayesian approximation~\cite{gal2016Dropout} is particularly appealing, as they have the same number of parameters as conventional NNs. However, the number of stochastic units required for their implementation is still a concern.




Moreover, in terms of hardware architectures, BayNNs are typically implemented on von Neumann architectures, where the memory and computation units are physically separated. Since BayNN applications are data intensive, data movement between the processor and memory becomes the bottleneck of the entire application, leading to the memory wall problem~\cite{merolla2014million, zou2021breaking}. Consequently, implementing BayNN to Computation-in-Memory (CIM) architectures with emerging resistive non-volatile memories (NVMs)~\cite{yu2018neuro} is an attractive option to reduce their inherent costs. In the CIM architecture, the common operation of NN, matrix-vector multiplication, is carried out inside the memory where data (NN parameters) already reside. Therefore, the memory-wall problem is alleviated, leading to a highly attractive solution to accelerate BayNN at the edge. 
However, implementing BayNN to the CIM architecture is not straightforward due to the deterministic nature of the architecture, the limited precision of the spintronic memories, and the design of stochastic units for Bayesian inference. 
Also, an approach that does not make changes to the conventional memory array but only changes to the peripheral circuitry is attractive.

{Therefore, in the literature, several methods~\cite{ahmed_spinbayes_2023, ahmed2023spindrop,soyed_TNANO23, soyed_nanoarch22, bonnet2023bringing, ahmed2023scalable, Dorrance2022} have been proposed to implement Dropout-based as well as other approximate BayNNs into CIM architectures with different NVM technologies. However, they have limitations, including a) requiring numerous Dropout modules (typically one for each neuron), b) not suitable for convolutional layers, and c) having high power consumption.}

{In this paper, we propose the \emph{Scale Dropout}, a novel regularization technique and \emph{Monte Carlo Scale Dropout (MC-Scale Dropout)} based BayNN for uncertainty estimation in Binary Neural Networks (BNNs)~\cite{hubara2016binarized}. BNNs utilize 1-bit precision weights and activations, effectively addressing the constraints posed by limited-precision spintronic memories. 
Moreover, by integrating Scale Dropout during inference, we achieve a robust uncertainty estimation, comparable to MC-Dropout. Our approach aims to provide a balanced trade-off between model uncertainty and the computational constraints of edge devices. The scale Dropout employs the proposed \emph{vector-wise} unitary Dropout technique, dropping out the entire scale vector, thus reducing memory and computational requirements without sacrificing the quality of uncertainty estimates. The scale vector, an additional parameter in BNN, is crucial in modern BNN algorithms to reduce quantization error. }

The primary contributions of our research are as follows:
\begin{itemize}
    \item We introduce a novel regularization technique named scale-Dropout, which can reduce co-adaptation in BNN training.
    \item We introduce the MC-Scale Dropout-based Bayesian approximation for efficient estimation in BNNs.
    \item We propose a novel CIM architecture where the model parameter is implemented with spintronic memories operating in the deterministic region and a spintronic-based scale Dropout module with the spintronic device operation in the stochastic regime. Our proposed CIM architecture does not imply changes to the common crossbar structure. It can reuse existing crossbar structures with only the peripheral circuitry modified for Bayesian inference.
\end{itemize}

Compared to existing work, our method requires \emph{only one Dropout module for the model, regardless of the size of the topology} leading to a highly scalable Bayesian approach for edge applications. The predictive performance of our method was extensively evaluated for various deep learning tasks, including classification and semantic segmentation, using different data sets and network topologies. Also, the effectiveness of our approach in estimating uncertainty is evaluated on various out-of-distribution data and metrics.

The structure of the paper is as follows: Section~\ref{sec:Preliminary} describes the background on Binary Neural Networks, traditional Dropout techniques, spintronic memories, CIM, and discusses related papers. In Section~\ref{sec:Proposed}, we present the proposed Scale Dropout technique, explaining its design, operation, and uncertainty estimation approach. In Section~\ref{sec:Hardware}, details of hardware implementation are discussed. In Section~\ref{sec:sim_setup}, simulation and experimental setup are discussed, and in Section~\ref{sec:Results}, our experimental results are presented at the algorithmic and hardware levels. Later, in Section~\ref{sec:discussion}, we discuss various aspects of the proposed method, and Section~\ref{sec:Conclusion} concludes the paper.

\section{Preliminaries}\label{sec:Preliminary}

\subsection{Binary NNs and Scaling}
BNNs have gained popularity as a model compression technique due to their ability to reduce memory requirements by $\sim 32\times$ compared to a full-precision NN. In BNNs, the weights and activations are binarized to $-1$ or $+1$ using the $\operatorname{sign}(.)$ function. Binarization further simplifies computationally intensive matrix-vector multiplication operations into computationally cheaper bitwise XNOR and bit counting operations, thus improving computational efficiency and leading to a speed-up of $\sim 58\times$ on CPU~\cite{rastegari2016xnor}.

{The scale vector $\mAlpha$ is a crucial aspect of BNN to alleviate the loss of accuracy due to binarization~\cite{rastegari2016xnor}. Here, a real-valued vector multiplies the weighted sum, weights, or activations of a layer. The scale vector can be defined in two ways, such as analytically calculated values that scale the binarized weights and activations for each layer in~\cite{rastegari2016xnor} or it can be learned through backpropagation similar to other parameters of the model~\cite{bulat2019xnor}. In terms of the location of the scale vector, applying the scale to the weight matrix of each layer before the XNOR operation (as done in~\cite{qin2020forward}) is possible in a CPU or GPU implementation but may not be as feasible for a spintronics-based CIM architecture.} This is because, depending on the shape of the scale vector, each neuron or channel will have a different scale factor, leading to different mapping strategy requirements for each neuron or channel. Similarly, input scaling is not feasible, as the inputs are directly converted to voltages and fed into the crossbar for computation.

In this paper, we specially design the scale vector and the application so that it can be implemented in the CIM architecture as well.

\subsection{Uncertainty Estimation}
Uncertainty estimation in neural networks is the process by which a model provides a measure of uncertainty or confidence along with its predictions. This measure typically captures the model's belief about the output, given the input data and the learned parameters~\cite{nguyen2015deep}. Uncertainty estimation enables NNs to express what they do not know. Traditional neural networks, despite their impressive prediction capabilities, fail to provide these crucial uncertainty estimates. 
There are two distinct types of uncertainty in deep learning: Aleatoric and epistemic~\cite{he2023survey}. Aleatoric or data uncertainty, which is irreducible even with more data, arises due to the inherent noise in the data generation process. On the other hand, uncertainty resulting from a lack of knowledge is referred to as epistemic or model uncertainty. In other words, it refers to the lack of knowledge of a model. Epistemic uncertainty can be eliminated with sufficient data from an unseen region or with knowledge about sources of uncertainty. Estimating epistemic uncertainty allows the NN to be aware of \emph{what they do not know}. Therefore, this paper focused on estimating the epistemic uncertainty of a model.

\subsection{Conventional Dropout Methods}
Dropout is a regularization technique commonly used in neural networks to prevent overfitting~\cite{srivastava2014Dropout}. During training, the Dropout randomly sets a proportion of the input units to 
\emph{zero} with a probability $p$ at each forward pass. It can be interpreted as training a large ensemble of "thinned" networks. The final prediction is then an ensemble prediction of these networks. However, Dropout is not applied during inference. 

There have been other variants of Dropout, e.g., Dropconnect proposed by Wan et al.~\cite{wan2013regularization}, which proposed to set individual weights and biases rather than neuron output to zero with some probability $p$. Spatial Dropout proposed by Tompson et al.~\cite{tompson2015efficient}, is a Dropout method targeted for convolutional neural networks. 
In Spatial Dropout, entire feature maps are dropped (set to zero) with probability $p$ rather than individual pixels to reduce spatial correlation. As a result, the network is prevented from using nearby pixels to recover information when a Dropout is applied.
{Variational Dropout, as proposed by Kingma et al.~\cite{kingma2015variational}, uses Gaussian multiplicative noise instead of zeroing out the weights or activations of the network. Their approach also enables the NN to automatically determine an effective Dropout probability for an entire network or for individual layers or neurons.
}

In this paper, we take a different approach to Dropout. Our approach is focused on the \emph{scale vectors} of a binary NN model. Additionally, our approach does not set the dropped value of the scale vector to zero, as done in existing work, as it prevents information flow. This means that the whole weighted sum of a layer would be zeroed out if the scaled vector were set to zero. Instead, our approach drops the entire scale vector to \emph{one}. That means in the dropped behavior, the scale vector is bypassed with a probability, and the weighted sum value of a layer remains unchanged.


\subsection{Bayesian Neural Networks}

A conventional NN represents a function \mbox{$\vnet: \mathbb{R}^D \times \mathbb{R}^P \to \mathbb{R}^C$} parameterized by the 
learnable parameters $\vtheta$. $D$, $P$, and $C$ represent the dimensions of inputs $\vx$, parameters $\vtheta$, and outputs $\vy$, respectively. Here, the parameter vector $\vtheta$ 
is a (point) estimate, that is, a single point value, and is found using a classic maximum likelihood estimation approach. In the maximum likelihood estimation method, 
a certain likelihood function $p(\vy \mid \vx, \vtheta)$ is maximized given the observed data $\mathcal{D}$ to obtain $\vtheta$ 
\cite{bishop2006pattern}. Despite the success of this approach, it inherently ignores the uncertainty in the estimation of the parameters $\vtheta$. 

In contrast, BayNNs are based on the Bayesian framework and offer an approach to uncertainty estimation. In a BayNN, the parameters $\vtheta$ of the network are treated as random variables. Consequently, a posterior distribution of the parameters $\theta$ is computed given the dataset $\mathcal{D}$.
Unfortunately, the exact computation of this posterior distribution $p(\vtheta | \mathcal{D})$, as well as the resulting posterior predictive distribution 
\begin{align}
p(\vy^* \mid \vx^*, \mathcal{D}) &= \int p(\vy^* \mid \vx^*, \vtheta) \, p(\vtheta \mid \mathcal{D}) \, \rm d \vtheta
\end{align}
is generally computationally intractable. 
To overcome this challenge, various approximation methods, such as Monte Carlo Dropout \cite{gal2016Dropout}, were developed to allow feasible approximations of uncertainty for BayNNs. The approximate posterior distribution is depicted as $q(y^*\mid x^*, \mathcal{D})$.

%
%
%
By introducing Dropout layers into the network during training and, more importantly, keeping them active during testing, MC-Dropout effectively simulates a sampling process from the posterior distribution of the parameters $\vtheta$. 
When the Dropout layers are kept active during testing, samples can be obtained from an approximate posterior.

This approach, while computationally cheaper than most other approximation techniques, provides a practical and efficient way to estimate predictive uncertainty, making BayNNs more accessible for practical applications \cite{gal2016Dropout}.

\subsection{Spintronic Memory Technology}

The primary component of STT-MRAM is the magnetic tunnel junction (MTJ). It comprises two ferromagnetic layers: the free layer and the reference layer, and is separated by a thin oxide layer. The magnetic orientation of the reference layer is fixed, but the orientation of the free layers can be changed (by passing a proper SET/RESET current) to be either parallel or anti-parallel, corresponding to low and high resistive states, respectively~\cite{dieny_opportunities_2020}. 
STT-MRAM offers several advantages, such as fast switching, high endurance, and CMOS compatibility \cite{lee_world-most_2022}. 

Despite their benefits, their resistance levels are low, typically within a few $k\Omega$. As a consequence, attempting to read all the bit cells in a crossbar array simultaneously can result in an excessively high current at the output of the crossbar. 

In order to address the challenge of integrating STT-MRAM for matrix-vector multiplication operation, alternative spintronic memory technologies can be considered, such as Spin-Orbit Torque (SOT) based MRAM (SOT-MRAM), a three-terminal device. In SOT-MRAM, the MTJ is mounted on a heavy-metal substrate. The resistance states of the SOT devices can be adjusted, allowing them to achieve resistance levels ranging up to several M$\Omega$~\cite{shao_roadmap_2021}. In addition, the reliability during read operation in SOT-MRAM is considerably enhanced, as they have separate read and write paths.

\subsection{Related Works}
Previous studies have investigated hardware solutions for Bayesian and Binary Neural Networks, which are usually based on CMOS technology.
In the paper~\cite{awano_bynqnet_2020}, a novel FPGA implementation is proposed that utilizes non-linear activation functions. However, this approach may be restricted when used with larger datasets. {With similar technology, work in~\cite{fan_high-performance_2021} proposed an architecture to implement MC-Dropout. The study described in~\cite{fan_fpga-based_2022} introduces an approach where only a partial BayNN is implemented, treating solely the last layers of the network as Bayesian.}
The technique described in \cite{malhotra_exploiting_2020} involves a CIM implementation in which the crossbar array stores the variance parameter and stochastic resistive RRAM devices are used to sample the probability distribution at the input of the array. This approach requires a single random element for each input, which is not very energy efficient.
In~\cite{dalgaty_situ_2021}, the authors take advantage of the non-idealities of RRAM devices to apply Bayesian learning.
In the paper by~\cite{soyed_nanoarch22}, an implementation of neuron Dropout was proposed that takes advantage of the stochastic and deterministic features of STT-MRAM. Unfortunately, this approach requires a random number generator (RNG) for each neuron, leading to a considerable increase in power consumption.
The research in \cite{bonnet2023bringing} showed the application of a set of resistive crossbar arrays to store probabilistic weights to execute BayNN. In \cite{ahmed_spinbayes_2023}, they proposed a suitable CIM architecture for BayNN, where multiple crossbar arrays are available and one is randomly chosen for each forward pass.
In Yang et al.~\cite{yang_all-spin_2020}, crossbar arrays were used to construct Bayesian neural networks with the help of low-barrier MTJs, resulting in a significant decrease in energy consumption. Despite the fact that memories with low-energy barriers are used, they have endurance limitations that can eventually have an impact on the precision of the CIM engine. The paper in~\cite{lu_algorithm-hardware_2022} presented an alternative implementation with MRAM-based crossbar arrays that can represent mean and variance. However, this approach required considerable pre-processing to encode the mean and variance in the crossbars.

In this paper, we propose an architecture that has a decreased dependence on RNGs. Also, we employ two arrays to store both the weights and the proposed Scale for a reduced overhead for BayNN implementation. 


\section{Proposed approach}\label{sec:Proposed}
\subsection{Scale Vector}
Considering the constraints and opportunities in the CIM architecture (see section{\ref{sec:Preliminary}}), we propose a hardware-software co-design approach for the scaling factor. Specifically, we design our scale factor (denoted as $\mAlpha$) to be learnable through a gradient descent algorithm and the same shape as the bias vector of a layer, $\boldsymbol{\mAlpha} \in \mathbb{R}^{\text{C}_\text{out}\times 1\times 1}$. Here, $\text{C}_\text{out}$ represents the number of output channels in convolutional layers and the number of neurons in linear layers. This choice is motivated by the desire to reduce memory overhead while ensuring compatibility with the CIM architecture. By making the scale factor learnable, we allow the training process to determine the optimal scale factor, making the model more adaptive and possibly improving its performance~\cite{bulat2019xnor}. Note that the weight and bias parameters and the $\mu$ and $\sigma$ variables of the batch normalization layer have the same value as the bias vector of a layer. Therefore, choosing the same shape of scale vector as those vectors leads to simplified computation and storage in CIM architecture.

\subsection{Scale Dropout Model Description}
Let a BNN with $L$ hidden layers and $\z^{(l-1)}$ denote the input vector, $\z^{(l)}$ denote the output vector, $\boldsymbol{\mAlpha}^{(l)}$ denote the scale vector, $\mW^{(l)}$ denote the weights and $\vb^{(l)}$ denote the biases of the layer $l$. The feed-forward operation (for $l = 0, \cdots, L-1$) of BNN can be described as

\begin{align}
\z^{(l)} &= (\operatorname{sign}(\mW^{(l)})^\top \otimes\operatorname{sign}(\z^{(l-1)}) + \vb^{(l)})\odot\boldsymbol{\mAlpha}^{(l)} \\
\hat{\z}^{(l)} &= \text{BatchNorm}_{\gamma, \beta} (\z^{(l)})\\
\va^{(l)} &= \phi(\hat{\z}^{(l)})
\end{align}

where $\phi$ denotes the element-wise nonlinear activation function for BNN, e.g., the Tanh function (hyperbolic Tangent), $\top$ denotes the matrix transpose operation and $\text{BatchNorm}_{\gamma, \beta} (\cdot)$ denotes the bach normalization~\cite{ioffe2015batch} with a learnable parameter $\gamma$ and $\beta$. In addition, $\odot$ denotes element-wise multiplication, and $\otimes$ denotes binary convolution. With Scale-Dropout, the feed-forward operation becomes :
\begin{align}\label{eq:scale_drop_feed}
d^{(l)} &\sim \text{Bernoulli}(p) \\
\hat{\boldsymbol{\mAlpha}}^{(l)} &= \boldsymbol{\mAlpha}^{(l)} \cdot d^{(l)} \\
\z^{(l)} &= (\operatorname{sign}(\mW^{(l)})^\top \otimes\operatorname{sign}(\z^{(l-1)}) + \vb^{(l)})\odot\hat{\mAlpha}^{(l)} \\
\hat{\z}^{(l)} &= \text{BatchNorm}_{\gamma, \beta} (\z^{(l)})\\
\va^{(l)} &= \phi(\hat{\z}^{(l)})
\end{align}

\begin{figure*}
    \centering
    \includegraphics[width=0.9\linewidth]{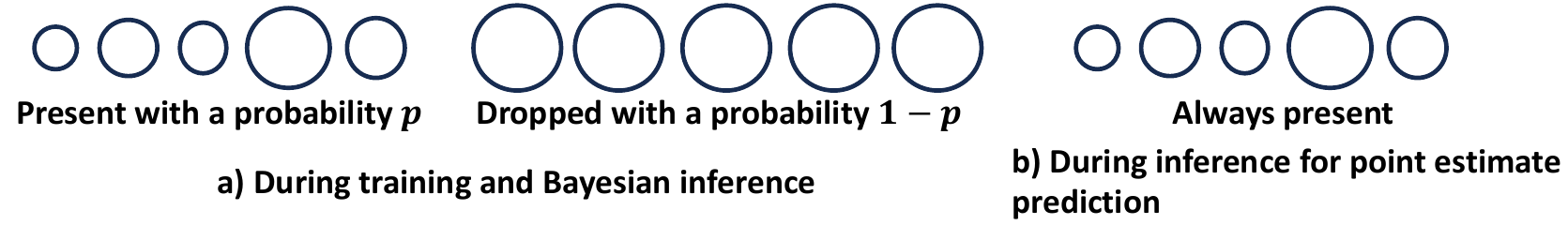}
    \caption{Several nodes (neurons) a) at training time that are scaled with a probability of $p$ and dropped (ignored) with a probability of $1-p$, b) At test time, if point estimate prediction is preferred, all the nodes are always scaled. However, for Bayesian inference, all nodes behave similarly to train time. Here, all the nodes are connected to the weights of the next layer after non-linear activation and Batch normalization, and their shapes represent scaling factors. All the dropped nodes have the same shape, indicating no scaling factor. 
    }
    \label{fig:scale_drop_schamatic}
\end{figure*}

Here, the Dropout mask for the scale Dropout is defined as a scalar $d \in \{0, 1\}$ and is independently sampled from a Bernoulli distribution with a probability parameter $p$ for each layer.

The scale vector multiplies the weighted sum of each layer. Therefore, if we were to set the scale values to zero (similar to traditional Dropout), it would lead to a complete loss of information in that layer.
To address this problem, we introduce an alternative approach called \emph{Unitary Dropout}. In this method, when the randomly generated Dropout mask is \emph{0}, all elements associated with the scale vector are set to \emph{1}.

As a result, during forward propagation, the network ignores the scale factors that correspond to the Dropout mask being \emph{0}, while the scaling factor retains its original value when the randomly generated Dropout mask is set to \emph{1}. Fig.~\ref{fig:scale_drop_schamatic} shows the scale Dropout concept during the train and inference time.

Although we have focused on \emph{Unitary Dropout} in this paper due to their simple implementation in the CIM architecture, other alternatives can also be considered. For instance, \emph{Average Scale Dropout} and \emph{Random Scale Dropout}. In \emph{Average Scale Dropout}, instead of setting the scale vector to \emph{one}, it involves dropping to the average of the scale vector, \mbox{$\Bar{\boldsymbol{\mAlpha}}=\frac{1}{\text{C}_\text{out}}\sum_{i=1}^{\text{C}_\text{out}}\boldsymbol{\mAlpha}_i$}. On the other hand, the Random Scale-Dropout method involves replacing the dropped scale with a random value sampled from a predefined distribution, for example, a uniform distribution. 

Additionally, to reduce the number of Dropout modules to one in the CIM architecture, the entire scale vector is dropped at the same time. However, the proposed Scale-Dropout can be applied to the scale vector element-wise at the cost of a large number of Dropout modules.

\subsection{Co-adaptation Mitigation}
The introduction of the proposed scale Dropout imposes randomness in the scale vector and, in turn, the activation of a layer. Thereby, it can potentially reduce co-adaptation between the scale vector and the binary weights. When $\boldsymbol{\mAlpha}$ is treated as a random variable during training, the model is less dependent on specific scale values, promoting a more diverse range of features in the BNN. This phenomenon can be expressed mathematically as increased variance in the learned representations across the network, thus reducing co-adaptation.

\subsection{Choosing Dropout Probability}\label{sec:Adap_drop}
To choose a Dropout probability of the scale Dropout, we propose a \emph{layer-dependent adaptive scale Dropout} method. 
Specifically, a Dropout probability of $10$\% or $20$\% is used on layers with a comparably smaller number of parameters, but a larger Dropout probability, e.g., $50$\% is used on layers with a larger number of parameters. Consequently, unlike works~\cite{soyed_nanoarch22, soyed_TNANO23}, where many different kinds of implementation with different locations for the Dropout layer need to be explored, our approach does not require such exploration, as the Scale-Dropout is applied to all the binary layers. Also, it is not necessary to explore the various Dropout rates. Consequently, our approach stands out as a more deployment-ready solution compared to related works.

\subsection{Learning with Scale-Dropout}
The proposed BayNN with \emph{Scale-Dropout} can be trained using stochastic gradient descent, similar to standard BNN using existing algorithms such as~\cite{hubara2016binarized, rastegari2016xnor}. The only difference is that for each forward pass during training, we sample a \emph{scaled network} by applying Scale-Dropout. The forward and backward propagation for each iteration is performed only on this \emph{scaled network}. The gradients for each parameter are averaged over the training instances of each mini-batch. The training objective combining a Bayesian approximation and \emph{Scale-Dropout} is discussed in Section~\ref{sec:MCScaleDropout}.

Although \emph{ Scale-Dropout} alone offers several benefits, using \emph{Scale-Dropout} in conjunction with common regularization techniques such as L2 regularization, learning rate scheduling, data augmentation, and momentum for the gradient descent algorithm further improves accuracy.

\section{Scale-Dropout as a Bayesian Approximation} \label{sec:MCScaleDropout}
As stated previously in Section~\ref{sec:Preliminary}, an NN with standard Dropout can be used as an approximate method of Bayesian inference. Gal et al.~\cite{gal2016Dropout}
showed that learning an NN with Dropout and L2 regularization is equivalent to a Gaussian process. The optimization objective of their approach, named MC-Dropout, is given by
\begin{equation}\label{eq: MC_Dropout}
\mathcal{L}(\boldsymbol{\vtheta})_{\text{MC-Dropout}} = \mathcal{L}(\vtheta, \mathcal{D}) + \lambda\sum_{l=1}^L  (||\vw^l||^2_2 + ||\vb^l||^2_2).
\end{equation}

In this paper, we propose \emph{Monte Calo (MC)-Scale Dropout} based Bayesian approximation that uses \emph{Scale-Dropout} in place of the standard Dropout for Bayesian inference. Our approach expands the MC-Dropout approaches~\cite{soyed_nanoarch22, gal2016Dropout}, for BNN and better efficiency with specific learning objectives. In the following section, the learning objective and how to obtain the model uncertainty for the \emph{(MC)-Scale Dropout} are discussed in detail.

\subsection{Learning Objective}

For the proposed \emph{Monte Calo (MC)-Scale Dropout} objective, we introduce a regularization function for the scales $\boldsymbol{\mAlpha}$. Specifically, we design a regularization function that encourages the scale factor to be positive to preserve the sign of the computed $\z^{(l)}$ of a layer $l$. Also, it encourages the scale factor to be centered around one, so it scales up or down the element of $\z$ based on their contribution to the loss. 


To achieve a Bayesian approximation, we use a similar approach to MC-Dropout.
However, in MC-Dropout, activations are dropped to zero, which inspires the L2 regularization to push the weights towards zero. On the contrary, in our \emph{ Unitary Dropout} approach, the scale factors are dropped to \emph{one}. This promotes a regularization effect that encourages the scale vector to center around one, a key distinction that aligns better with the nature of binary networks where weights are binarized to $-1$ or $1$. The regularization function can be mathematically described by
\begin{equation}
    \varphi\sum_{l=1}^L  (1-\mu^l_{\mAlpha})^2.
\end{equation}
Here, $\mu_{\mAlpha}^l$ is the mean of the scale vector of a layer $l$ and $\varphi$ is the hyperparameter for controlling the strength of the regularization.

Despite the regularization of the scales, we also optionally apply the L2 regularization to the weights. Applying L2 regularization is a challenge in BNN. In BNN, real-valued proxy weights are binarized to $+1$ or $-1$, therefore, applying L2 regularization to either of them may not be beneficial~\cite{soyed_nanoarch22}. 

However, L2 regularization can be implemented in the actual real-valued weights, with binarization applied to the normalized weights within the output channel dimensions~\cite{soyed_TNANO23}. Opting for channel-wise normalization also proves advantageous in reducing binarization errors~\cite{qin2020forward}. To achieve this, the channel-wise mean is first computed:
\begin{equation}
\mu_{c} = \frac{1}{H \times \mathcal{W}} \sum_{h=1}^{H} \sum_{w=1}^{\mathcal{W}} \vw.
\end{equation}
Here, $H$ and $\mathcal{W}$ represent the height and width of the kernels in the weight matrix, the last two dimensions of $\mathcal{W}$. Subsequently, the channel-wise mean $\mu_{c}$ is subtracted from the proxy weights (real-valued):
\begin{equation}
\hat{\vw} = \vw-\mu_{c}.
\end{equation}
Following that, the channel-wise standard deviation is calculated on zero-centered weights.
\begin{equation}
\sigma_{c}^2 =\frac{1}{H \times \mathcal{W}} \sum_{h=1}^{H} \sum_{w=1}^{\mathcal{W}} \vw^2- \hat{\vw}^2.
\end{equation}
Note that $\sigma_{c}^2$ calculation is simplified for efficiency reasons. Lastly, the channel-wise standard deviation divides the zero centered weight for channel-wise normalization as:
\begin{equation}
\Tilde{\vw} = \frac{\hat{\vw}}{\sigma_{c}}.
\end{equation}

Consequently, binarization on the channel-wise normalized weights can be defined as:
\begin{equation}
\vw^* = 
\begin{cases} 
+1 & \text{if } \Tilde{\vw} \geq 0 \\
-1 & \text{otherwise}
\end{cases}
\end{equation}
Note that, channel-wise weight normalization has become standard practice in modern BNN models.

The overall objective of the MC-Scale Dropout with both scales and weight Dropout is defined as:

\begin{equation} \label{eq:MC-scaleDropout}
\mathcal{L}({\vtheta})_{\text{MC-Scale Dropout}} = \mathcal{L}(\vtheta, \mathcal{D}) + \lambda\sum_{l=1}^L  ||\mW^l||^2_2 + \varphi\sum_{l=1}^L  (1-\mu^l_{\mAlpha})^2.
\end{equation}

Here, $\lambda$ is the weight decay hyperparameter of the weight regularization.

\subsection{Obtaining Model Uncertainty}

To obtain the uncertainty of the model, we perform $T$ forward passes with the proposed \emph{Scale-Dropout} enabled during Bayesian inference. During each of the $T$ forward passes, we sample an independent and identically distributed random Dropout mask from the Bernoulli distribution for each layer $\{d^{(l)}_t,\cdots,d^{(L)}_t\}^T_{t=1}$, giving $T$ stochastic scale vectors $\{\mAlpha^{(l)}_t,\cdots,\mAlpha^{(L)}_t\}^T_{t=1}$ and ultimately stochastic weighted sums $\{\z^{(l)}_t,\cdots,\z^{(L)}_t\}^T_{t=1}$.
The predictive mean is given by:
\begin{equation}\label{eq:pred_mean}
    E_{q(y^*\mid x^*, \mathcal{D})}(y^*) \approx \frac{1}{T} \sum_{t=1}^T \Tilde{y}^*_t (x^*, \z^{(l)}_t,\cdots,\z^{(L)}_t)
\end{equation}
Here, $x^*$ is the test input, $q(y^*\mid x^*, \mathcal{D})$ is the posterior distribution, $\Tilde{y}$ is the stochastic prediction, and $y^*$ is the final prediction.
We refer to this Monte Carlo estimate as the \emph{MC-Scale Dropout}. In practice, this is equivalent to performing $T$ stochastic forward passes through the network and averaging the results. In the literature, this is known as model averaging~\cite{gal2016Dropout}.

{In terms of the posterior distribution of the output, equations~\ref{eq:scale_drop_feed} can be modified as 
\begin{align}
\z^{(l)} &= S^{(l)}\odot \text{diag}(d) \\
d^{(l)} &\sim \text{Bernoulli}(p) \text{ for } l=1, \cdots, L
\end{align}
Here, $S$ represents the weighted sum of a layer. Batch normalization is applied to $\z^{(l)}$. Thus, the sampling process of the Dropout mask is the same as that of the MC-Dropout. In an empirical evaluation (shown later in the~\ref{sec:discussion}), we observed that the distribution of output for each class approaches a Gaussian distribution as the number of stochastic forward passes (T) through the network increases. This is due to the aggregate effect of the scalar dropout mask over many forward passes, which can be seen as introducing a form of multiplicative noise.}



Uncertainty estimates of the prediction can be obtained from the variance of the $T$ forward passes as
\begin{multline}\label{eq:pred_var}
    \text{Var}_{q(y^*\mid x^*, \mathcal{D})}(y^*) \\ \approx \frac{1}{T} \sum_{t=1}^T (\Tilde{y}^*_t (x^*, \z^{(l)}_t,\cdots,\z^{(L)}_t) - E_{q(y^*\mid x^*, \mathcal{D})}(y^*))^2
\end{multline}
In addition, the $K$\% confidence interval (CI) can also be used as an uncertainty estimate of the MC-Scale Dropout model. According to the central limit Theorem, for sufficiently large $T$,  \{$\Tilde{y}^*_1\cdots\Tilde{y}^*_T$\} follow a normal distribution. For a $K$\% confidence interval, we use the percentiles of the predictions. Let $Q_{\eta/2}$ be the $\eta/2$ quantile of the predictions. Where $\eta = 1 - K/100$. Consequently, the $K$\% confidence interval is given by
\begin{equation}\label{eq:CI}
    \text{CI} = \left[\mu_y - Q_{\eta/2} \frac{\sigma_y}{\sqrt{T}}, \mu_y + Q_{\eta/2} \frac{\sigma_y}{\sqrt{T}}\right].
\end{equation}
Here, $\mu_Y$ and $\sigma_Y$ represent predictive mean $E_{q(y^*\mid x^*, \mathcal{D})}(y^*)$, and variance $\text{Var}_{q(y^*\mid x^*, \mathcal{D})}(y^*)$ from formulas \ref{eq:pred_mean} and \ref{eq:pred_var}, respectively.
For sufficiently large $T$, the confidence interval can be approximated by directly calculating the $\frac{100-k}{2}$ and $\frac{100+k}{2}$ quantile (for a $K$\% CI) of the predictions as 
\begin{equation}\label{eq:CI_approx}
    \text{CI} \approx \left[\text{percentile}\left(\frac{100-k}{2}\right), \text{percentile}\left(\frac{100+k}{2}\right)\right].
\end{equation}

\section{Hardware Implementation}\label{sec:Hardware}

\subsection{Modelling Spintronic-based Scale Dropout}\label{sec:spin_scale_drop_model}
In our design, only \emph{one} spintronic-based Dropout (namd here Spin-ScaleDrop) module is designed and implemented for the entire neural network. 
Thus, the proposed Spin-ScaleDrop module is reused for all layers of the CIM architecture. After the computation of a layer is performed, a new Dropout mask from the Spin-ScaleDrop Module module is sampled for the next layer.
However, due to the manufacturing and infield variation of the MTJs in the Spin-ScaleDrop, the Dropout probability itself becomes a stochastic variable. We model the variation as a Gaussian distribution, the mean of the distribution $\mu$ represents the expected Dropout probability, and $\sigma$ represents the device variations. Therefore, the probability of Dropout $p$ of a layer $l$ can be modelled as
\begin{equation}\label{eq:spin_drop}
\hat{p}_l = p_l + \epsilon \quad \text{with} \quad \epsilon \sim \mathcal{N}(\mu,\sigma^{2}).
\end{equation}

Here, $\hat{p}_l$ denotes the probability of Dropout with variation in the process. The feed-forward operation expressed in equation~\ref{eq:scale_drop_feed} remains the same, with only $\hat{p}_l$ used as the Dropout probability. Note that a probability has to be in $[0,1]$, since $p$ is usually chosen between $0.1$ and $0.5$, it is unlikely that $p$ crosses this range due to variation. The Dropout probability typically varies from $3\%$ to $10\%$.

\subsection{Designing Spintronic-Based Scale Dropout Module}

\begin{figure}
    \centering
    \includegraphics[width=0.6\columnwidth]{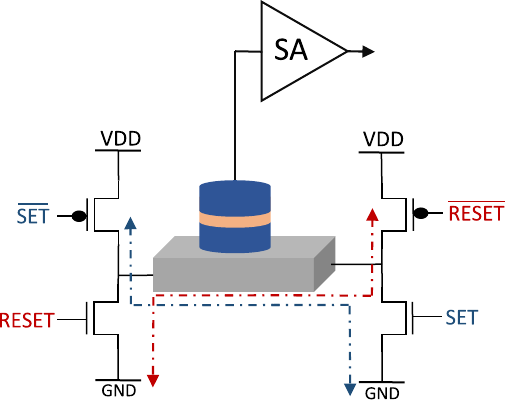}
    \caption{Spin Scale-Dropout Module based on SOT MTJ.}
    \label{fig:spin_RNG}
\end{figure}

The Spin-ScaleDrop module is designed by harnessing the stochastic regime of an MTJ and is utilized as a random number generator. The probability density function governing the switching of the SOT-MTJ follows an exponential distribution and is expressed as~\cite{lee_emerging_2016}:
\begin{subequations}
\label{eq:psw}
    \begin{equation}
    \label{eq:psw1}
        p_{\rm sw} = 1-\exp\left(\frac{t}{\tau}\right)
    \end{equation}

    \begin{equation}
    \label{eq:psw2}
        \tau = \tau_0 \exp\left[\frac{\Delta E}{k_B \mathcal{T}}\left({1-2\frac{I}{I_{c0}}(\frac{\pi}{2} - \frac{I}{I_{c0}}})\right)\right]
    \end{equation}
\end{subequations}

Here, $\Delta E$ is the thermal stability factor, $I$ is the applied current through the SOT-track, $t$ is the pulse duration, $\tau_0$ is the attempt time, $I_{c0}$ is the critical current at \SI{0}{\kelvin}, $k_B$ is the Boltzmann constant and $\mathcal{T}$ is the temperature. $I_{c0}$ represents the minimum current required to switch the MTJ. 
The equation~\eqref{eq:psw} is used to model the switching behavior of the SOT-MTJ for different switching currents while keeping the pulse width fixed at \SI{10}{\nano\second}. 
To generate the bidirectional current across the SOT track, four transistors are added, as shown in Fig.~\ref{fig:spin_RNG}. The desired switching probability of 50\% is achieved by programming the MTJs through successive "SET" and "RESET" operations.

To ensure reliable MTJ switching, the write duration is set to \SI{10}{\nano\second} for the SET operation and to \SI{5}{\nano\second} for the RESET operation. The state of the MTJ is read using a Sense Amplifier (SA, in Fig.\ref{fig:spin_RNG}). The SET and RESET cycles are repeated to generate a stochastic sequence. 
The Scale Dropout Module allows for the stochastic activation of the Scale vector that is stored in the neighboring memory. 


\subsection{Proposed Spintronics-based CIM Architecture}

\begin{figure}
    \centering
    \includegraphics[width=0.8\columnwidth]{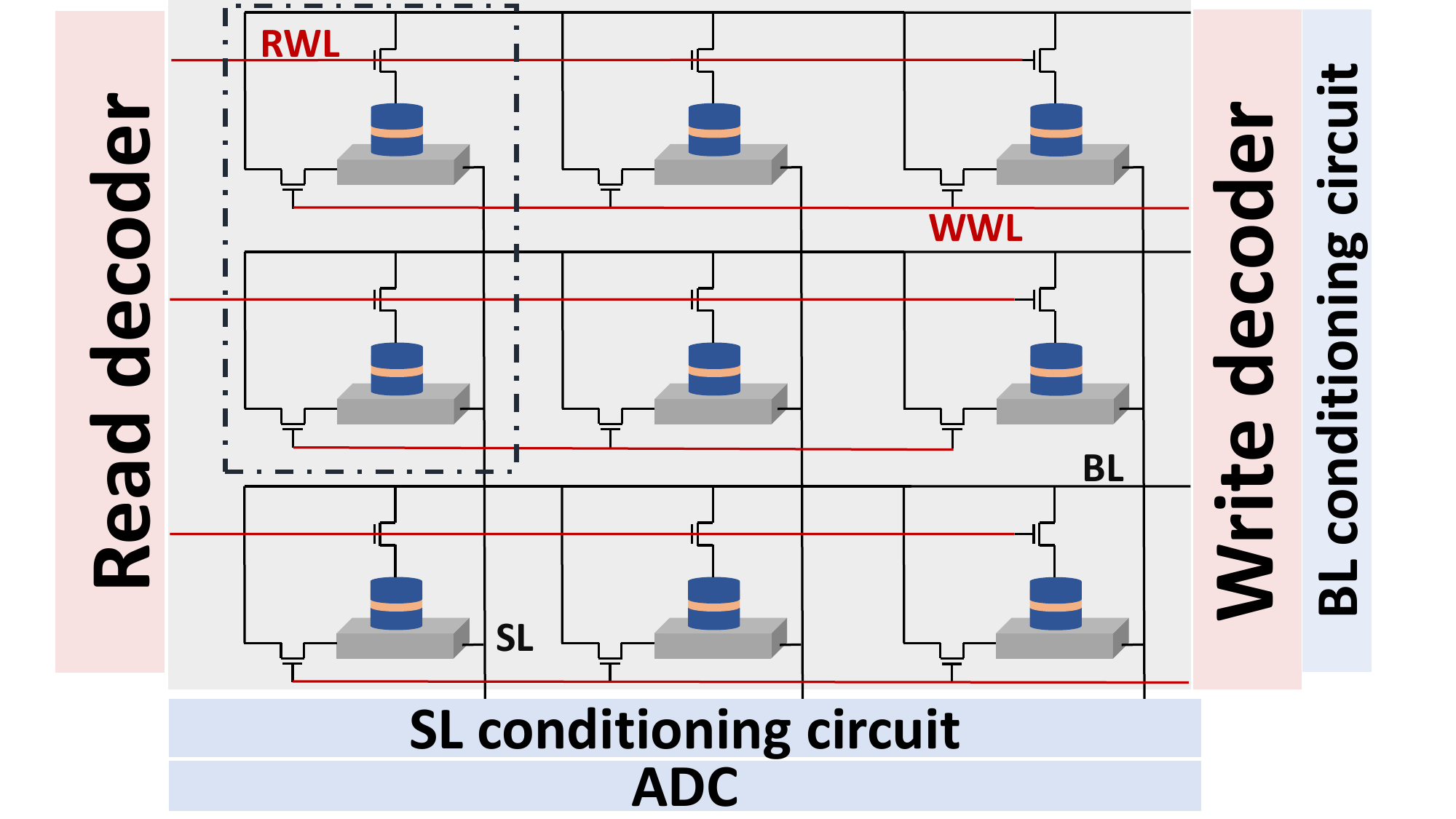}
    \caption{Binary SOT crossbar array for the Bayesian inference.}
    \label{fig:crossbar}
\end{figure}

\begin{figure}
   \centering
   \includegraphics[width=0.8\columnwidth]{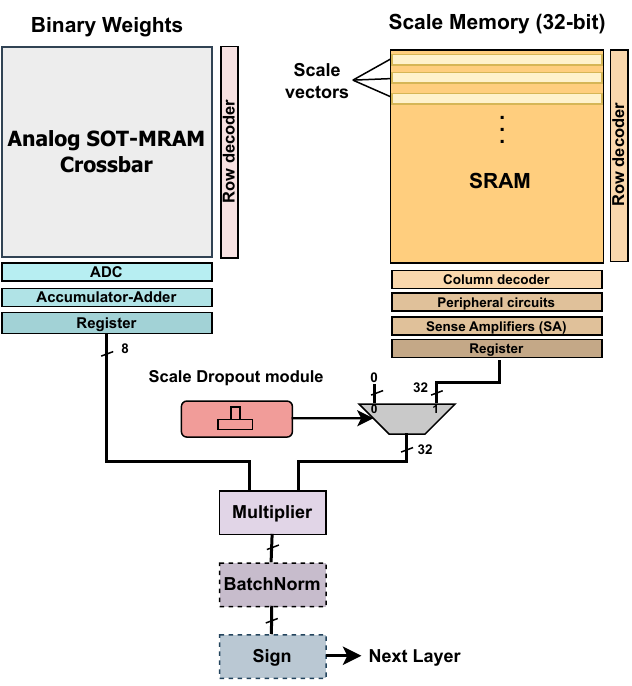}
   \caption{Proposed inference architecture for Scale-Dropout.}
   \label{fig:architecture}
\end{figure}

In spintronic-based CIM architectures, the SOT-MRAM devices are arranged in a crossbar fashion, with an MRAM device at each crosspoint (see Fig.~\ref{fig:crossbar}). For inference, the mapping of the trained binary weights to the array is performed with a one-time write operation. In BNN, XNOR and the bit-counting operation are performed instead of the weighted sum operation~\cite{hubara2016binarized}. The XNOR operation in CIM is shown in Table~\ref{tab:xnor_table} and the encoding of the respective $+1$ and $-1$ weights with the complementary bit cell is shown in Fig.~\ref{fig:crossbar}. 

\begin{table}
    \centering
    
    \caption{Showing Xnor operation for BNN implementation.}
    \label{tab:xnor_table}
    \resizebox{\linewidth}{!}{
    
    \begin{tabular}{|c|c|c|c|}
    \hline 
    Input/Activation & weight & Bitwise XNOR & Effective Resistance\tabularnewline
    \hline 
    \hline 
    $+1$ & $+1$ & \multirow{2}{*}{+1} & \multirow{2}{*}{LRS}\tabularnewline
    \cline{1-2} 
    $-1$ & $-1$ &  & \tabularnewline
    \hline 
    $+1$ & $-1$ & \multirow{2}{*}{-1} & \multirow{2}{*}{HRS}\tabularnewline
    \cline{1-2} 
    $-1$ & $+1$ &  & \tabularnewline
    \hline 
    \end{tabular}
    }
\end{table}

The mapping of the weight matrix of various NN layers to the crossbar arrays is challenging because of their varying shapes and dimensions. While mapping Fully Connected (FC) layers is relatively simple due to their 2D weight matrices ($\mathbb{R}^{m \times n}$), Convolutional (conv) layers pose challenges due to their 4D structures ($\mathbb{R}^{K \times K \times C_{\rm in} \times C_{\rm out}}$), with $K$ denoting the kernel size, $C_{\rm in}$ and $C_{\rm out}$  denoting the number of input and output channels, respectively. 

There are primarily two strategies for mapping convolutional layers. The first strategy \circled{1} involves unrolling each kernel of shape $K\times K\times C_{in}$ into a column of the crossbar~\cite{gokmen_training_2017}. The second strategy \circled{2} maps each kernel to multiple smaller crossbars of shape $C_{in}\times C_{out}$, arranged in a $K\times K$ layout~\cite{peng_optimizing_2019}.

During inference (online operation), each element of the binary input vector $x$ for a layer is converted into a ($0$, $1$) or ($1$, $0$) signal and fed into the crossbar array for inference. This architecture allows for parallel computation and outputs the weighted sum results as currents flow through each source line. Finally, the analog currents are converted to digital signals using Analog-to-Digital Converters (ADCs) and passed on to the Accumulator-Adder module to sum up the partial matrix-vectors multiplication. These partial multiplications are then stored in registers and multiplied with the Scale memory.  

Regarding the scale vectors, they are stored in a nearby 32-bit SRAM memory. In the scale memory, each row stores a scale vector. The column dimension of the SRAM memory depends on the maximum number of neurons or channels within the NN layers, and the row dimension depends on the number of layers in the model. This scale vector is subsequently applied, depending on the stochastic activation by the Scale Dropout module, using a multiplexer.

Recent state-of-the-art CNN topologies, e.g., ResNet and DenseNet, use skip connections. In Computing in Memory architectures, skip-connection can be implemented by selectively routing the output signals through the crossbars and summing them with digital circuits. Since layer-by-layer computations are sequential, signals for these connections can be stored in a buffer memory until the computation of the following layers is completed.

\section{Experimental and Simulation Setup}\label{sec:sim_setup}

\subsection{Datasets}
\subsubsection{In distribution (ID) Dataset}

To evaluate both predictive performance and uncertainty estimation, we have used several challenging benchmark and real-world biomedical in-distribution datasets on various learning paradigms (classification and semantic segmentation) in the context of Bayesian deep learning. An in-distribution dataset refers to a set of data samples that come from the same distribution as the data the model was trained on. For example, if a model is trained on images of an aeroplane, automobile, bird, cat, deer, dog, frog, horse, ship, and truck from CIFAR-10, then, during inference, more images of CIFAR-10 (although not seen during training) would be considered an in-distribution dataset. 

Specifically, for classification, we have used the CIFAR-10. 
Furthermore, 
for biomedical semantic segmentation, breast ultrasound scans (for breast cancer)\cite{al2020dataset}, COVID-19 lung computed tomography (CT)~\cite{ma2020covid}, and skin cancer~\cite{mendoncca2013ph}. The breast cancer dataset containing ultrasound scans is a vital resource used for the early detection of breast cancer, one of the leading causes of death among women worldwide. The dataset is classified into three classes, normal, benign, and malignant images, and
has a total of $780$ images with a size of $500 \times 500$ pixels on average. On the other hand, the Skin Cancer dataset for Biomedical Segmentation contains $200$ dermoscopic images of shape $572\times 765$ pixels with their corresponding label masks. 
Accurate prediction of skin cancer allows computer-aided diagnostic systems to assist medical professionals in the early detection and precise delineation of skin lesions.
Lastly, the COVID-19 lung CT dataset contains anonymized human lung CT scans with different levels of severity in COVID disease. 

Evaluating the proposed method on various datasets shows the scalability and generality of our proposed approach. Note that semantic segmentation, which involves segmenting an image into multiple sections and labeling each pixel with its corresponding class label, is regarded as more difficult than classification tasks due to its finer granularity.

We have applied random data augmentation and dataset normalization on all the datasets during training to improve accuracy. For example, for CIFAR-10 datasets, we have applied RandomHorizontalFlip and RandomResizedCrop type random data augmentation.

\subsubsection{OOD Dataset}
We used six additional OOD datasets to evaluate the efficacy of our method in dealing with data uncertainty. :
\begin{enumerate*}
  \item Gaussian noise ($\hat{\mathcal{D}}_1$): Each pixel of the image is generated by sampling random noise from a unit Gaussian distribution, $\x\sim \mathcal{N}(0,1)$,
  \item Uniform noise ($\hat{\mathcal{D}}_2$): Each pixel of the image is generated by sampling random noise from a uniform distribution,  $\x \sim \mathcal{U}(0,1)$,
  \item CIFAR-10 with Gaussian noise ($\hat{\mathcal{D}}_3$): Each pixel of the CIFAR-10 images is corrupted with Gaussian noise,
  \item CIFAR-10 with uniform noise ($\hat{\mathcal{D}}_4$): Each pixel of the CIFAR-10 images is corrupted with uniform noise,
  \item SVHN: Google street view house numbers dataset~\cite{netzer2011reading}, and
  \item STL10: a dataset containing images from the popular ImageNet dataset~\cite{coates2011analysis}.
\end{enumerate*}
Each of these OOD datasets contains $8000$ images, and the images have the same dimensions as the original CIFAR-10 dataset ($32 \times 32$ pixels).

\subsection{Evaluated Topologies and Training setting}

The proposed Scale-Dropout is evaluated for its predictive performance and uncertainty estimation in state-of-the-art convolutional NN (CNN) topologies, including ResNet~\cite{he2016deep}, and VGG~\cite{simonyan2014very} for benchmark classification tasks. In the case of biomedical image segmentation tasks, U-Net~\cite{ronneberger2015u}, and Bayesian SegNet~\cite{BMVC2017_57}, topologies are used. The U-Net topology consists of a contracting path and an expansive path with skip connections, which gives it the U-shaped architecture. On the other hand, Bayesian SegNet is a deep convolutional encoder-decoder architecture for semantic image segmentation. 

All models are trained with the Adam optimization algorithm with default settings in the PyTorch framework to minimize the proposed objective function with a weight decay rate of $\lambda=1\times 10^{-5}$ and $\varphi =1\times10^{-5}$. 
Classification and segmentation tasks are trained for 300 epochs.

All weights and activations of models for classification tasks are binarized (1-bit model). The activations of biomedical semantic segmentation models are quantized to 4 bits, but their weights are kept binary. As stated previously, semantic segmentation tasks are more difficult and therefore, they require slightly more bit precision at the activation for accurate predictions. We have used the activation quantization algorithm proposed in~\cite{choi2018pact} to quantize the activations to 4 bits. Since 1-bit weights are still maintained, the crossbar structure does not need to be modified. In fact, only peripheral modifications, such as ADC with increased bit resolution are required.

We have used the recently proposed IrNet~\cite{qin2020forward} binarization algorithm to implement the proposed learnable scale and Scale-Dropout. Note that (to our knowledge) any binarization algorithm can be extended with our method with slight modification, i.e., add a learnable scale vector and scale Dropout.

\subsection{Evaluation Metrics}

The evaluation metrics utilized to assess the effectiveness of segmentation tasks include pixel-wise accuracy, Intersection-Over-Union (IoU), Sensitivity, Specificity, Area Under the ROC Curve (AUC), F1 Score, and precision. Among these metrics, IoU holds particular significance, as it comprises the ratio between the area of overlap and the area of union between the predicted and ground-truth segments. Pixel-wise accuracy, on the other hand, quantifies the percentage of pixels in the predicted image that have been correctly classified. Specificity conveys the model's capability to accurately recognize actual negative cases, while Sensitivity reflects its ability to correctly identify actual positive cases. AUC is a single-number summary of the true-positive and false-positive rates, with higher values indicating superior performance. The F1 score serves as a comprehensive metric by integrating both precision and recall to measure the accuracy of a model on a dataset. Lastly, precision indicates the proportion of positive predictions that are genuinely correct.

On the other hand, classification tasks are evaluated for their inference accuracy, which constitutes the ratio of images correctly classified by the model over the total number of images in the validation dataset.

(Epistemic) Uncertainty estimation of the models is evaluated on predictive variations, entropy, and confidence interval with $K=95\%$ based on Equations~\ref{eq:pred_var},\ref{eq:CI_approx}, respectively. Out-of-distribution data is detected as:
\begin{equation}
\begin{cases} 
\text{OOD}, & \text{if } \max\left(\mathcal{Q}\left(\{ \hat{\y_t}^*\}_{t=1}^T\right)\right) < 0.95 \\
\text{ID}, & \text{otherwise}.
\end{cases}
\end{equation}

Here, $\hat{\y_t}^*$ represents the softmax output obtained during the stochastic forward pass of the MC run $t$ out of the $T$ runs. The function $\mathcal{Q}(\cdot)$ calculates the 1-th quantiles among the set of values, while $\max(\cdot)$ finds the maximum confidence score among output classes. The classification as ID or OOD depends on whether the maximum value from the 10th percentile of the averaged outputs is less than 0.95 (for OOD) or not (for ID). The underlying idea of our OOD detection is that for in-distribution data, most confidence scores from the $T$ MC runs are high and close to one another, resulting in low variance. In contrast, for out-of-distribution data, confidence scores exhibit higher variance.

\subsection{Architectural Simulation}
To carry out the architectural simulation, we first obtained the circuit specifications for the peripheral blocks, as outlined in Section \ref{sec:Hardware}. We then independently simulated each component of the architecture to gauge its energy utilization. Both the crossbar array and the Spin Scale-Dropout module were analyzed using an electrical simulator such as the Simulation Program for Integrated Circuit (SPICE), to assess their energy consumption. The use of high-resistance SOT devices~\cite{danouchi_spin_2022}, in conjunction with the binary nature of the network, serves to reduce the overhead related to peripheral elements.

The Accumulator-Adder, Comparator, and Averaging circuits were synthesized using the Synopsys Design Compiler, leveraging the TSMC 40 nm low-power Process Design Kit (PDK). For the CIM operation, decoding and sensing were assessed at the circuit-array level using NVsim (NonVolatile memory simulator)~\cite{nvsim6218223}. To achieve this, we modified the NVsim simulator to accommodate multiple active cells, thus simulating CIM operation accurately. Additionally, we substituted the single-bit sense amplifiers with multi-bit ADCs. Performance metrics for each discrete component are shown in Table \ref{tab:circuit_data}.

\begin{table}
\footnotesize
 \centering
    \caption{Energy estimation for the different elements of the architecture for one reading operation.}
   
    \resizebox{\linewidth}{!}{
    \begin{tabular}{|c|c|c|c|}
    \hline
    Circuit                   & Energy  & Circuit           & Energy \\ \hline
    Memory (Decoding/Sensing) & \SI{4.76}{\pico\joule}                   & Adder-Accumulator & \SI{0.12}{\pico\joule}                   \\ \hline
    Spintronic RNG        & \SI{3.80}{\pico\joule}                    & Comparator        & \SI{0.01}{\pico\joule}                    \\ \hline
    Averaging block              & \SI{18.42}{\pico\joule}                &      Crossbar  array        &  \SI{0.65}{\pico\joule}                       \\ \hline
\end{tabular}}
    \label{tab:circuit_data}
\end{table}


\section{Evaluation}\label{sec:Results}
\label{sec:Predictive Performance}
\begin{table}
\centering
\caption{Predictive performance of the proposed MC-SparialDropout method in comparison with SOTA methods on CIFAR-10. The accuracy closest to the MC-Dropout is in bold, and the number in the bracket shows the standard deviations of the accuracy after different repetitions.
}
\resizebox{\linewidth}{!}{
\begin{tabular}{|c|c|c|c|c|}
\hline
Topology                   & Method            & Bit-width (W/A) & Bayesian & Inference Accuracy \\ \hline
\multirow{5}{*}{ResNet-18} & FP                & 32/32           &       No   & $93.0\%$           \\ \cline{2-5} 
                           & RAD~\cite{rad}               & 1/1             &     No     & $90.5\%$           \\ \cline{2-5} 
                           & IR-Net~\cite{qin2020forward}            & 1/1             &     No     & $91.5\%$           \\ \cline{2-5}
                           &  MC-Dropout(HTanh)~\cite{gal2016Dropout}            & 32/32             &     Yes     & {$90.56\%$}           \\ \cline{2-5} 
                           & SpinDrop~\cite{soyed_nanoarch22}          & 1/1             &     Yes     & $90.48\%$              \\ \cline{2-5} 
                           & Spatial-SpinDrop~\cite{soyed_TNANO23}          & 1/1             &     Yes     & $91.34\%$              \\ \cline{2-5}
                           & \textbf{Proposed} & \textbf{1/1}             &     \textbf{Yes}     & $\mathbf{91.52\%(\pm 0.047)}$              \\ \hline
\multirow{6}{*}{ResNet-20} & FP                & 32/32           &       No   & $91.7\%$           \\ \cline{2-5} 
                           & DoReFa~\cite{dorefa}            & 1/1             &     No     & $79.3\%$           \\ \cline{2-5} 
                           & DSQ~\cite{dsq}               & 1/1             &     No     & $84.1\%$           \\ \cline{2-5} 
                           & IR-Net~\cite{qin2020forward}            & 1/1             &     No     & $85.4\%$           \\ \cline{2-5}
                           & MC-Dropout(HTanh)~\cite{gal2016Dropout}            & 32/32             &     Yes     & $86.94\%$           \\ \cline{2-5}
                           & Spatial-SpinDrop~\cite{soyed_TNANO23}            & 1/1             &     Yes     & $84.71\%$           \\ \cline{2-5}
                           & \textbf{Proposed} & \textbf{1/1}             &     \textbf{Yes}     & $\mathbf{86.04\% (\pm 0.039)}$            \\ \hline
\multirow{8}{*}{VGG}       & FP                & 32/32           &       No   & $91.7\%$           \\ \cline{2-5} 
                           & LAB~\cite{LAB}               & 1/1             &     No     & $87.7\%$           \\ \cline{2-5} 
                           & XNOR~\cite{rastegari2016xnor}              & 1/1             &     No     & $89.8\%$           \\ \cline{2-5} 
                           & BNN~\cite{hubara2016binarized}               & 1/1             &     No     & $89.9\%$           \\ \cline{2-5} 
                           & RAD~\cite{rad}               & 1/1             &     No     & $90.0\%$           \\ \cline{2-5} 
                           & IR-Net~\cite{qin2020forward}            & 1/1             &     No     & $90.4\%$           \\ \cline{2-5} 
                           & {MC-Dropout (HTanH)}~\cite{gal2016Dropout}            & {32/32   }          &     Yes     & $89.49\%$           \\ \cline{2-5}
                           & MC-Dropout (ReLU)~\cite{gal2016Dropout}            & 32/32             &     Yes     & $91.64\%$           \\ \cline{2-5}
                           & MC-DropConnect~\cite{mobiny2021dropconnect}            & 32/32             &     Yes     & $91.36\%$           \\ \cline{2-5}
                           & \textbf{SpinDrop~\cite{soyed_nanoarch22}}          & \textbf{1/1}             &     \textbf{Yes}     & $\mathbf{91.95\%}$            \\ \cline{2-5} 
                           & Spatial-SpinDrop~\cite{soyed_TNANO23}          & 1/1             &     Yes     & $90.34\%$            \\ \cline{2-5} 
                           & {Proposed} & 1/1             &     Yes     & ${90.45\%(\pm 0.052)}$            \\ \hline
\end{tabular}
}
\label{tab:comp_mc_drop}
\end{table}

\subsection{Predictive Performance}

\subsubsection{Comparison With State-of-the-Art Algorithms}

The predictive performance of our method is comparable to the SOTA binary Bayesian NN methods, as shown in Table~\ref{tab:comp_mc_drop} on a range of CNN architectures, including VGG, ResNet-18, and ResNet-20, evaluated on the CIFAR-10 dataset. In the worst case, the predictive performance is $1.45\%$ below the SpinDrop~\cite{soyed_nanoarch22} method for the VGG topology. Here, we assumed that there are no device variations in the spintronics-based scale Dropout module. For a fair comparison, we used the same network size as those used in their work. However, the hardware implementation of our solution may lead to a smaller area and a better power-performance product owing to a simpler spintronics-based Dropout module design. In our analysis, we have used layer-dependent adaptive Dropout rates
(See Section~\ref{sec:Adap_drop})
to scale Dropout. {The low variance in inference accuracy (numbers in brackets) shows the stability of the proposed approach.}

{In terms of the activation function, the proposed binary BayNN uses the \textit{Sign(.)} function, which is an approximation of the hard Tanh function. In this case, the proposed method performs similarly to the MC-Dropout method. However, in the case of the ReLU activation function in the MC-Dropout model, the accuracy of the MC-Dropout model increases. Thus, the difference between the proposed method and the MC-Dropout increases to around $\sim 1\%$.}

Furthermore, our proposed method achieves an improvement of up to $6.74\%$ in inference accuracy compared to the SOTA point estimate BNN algorithm.
However, since our method is built on top of the IR-Net BNN algorithm~\cite{qin2020forward}, predictive performance should be comparable to their approach. As depicted in Table~\ref{tab:comp_mc_drop}, the predictive performance is in the worst case $0.18\%$ lower, which is negligible. Similarly, accuracy is comparable to the full precision model, depicting that our method in general does not increase quantization error. Note that in the full precision model, the ReLU function is used as the activation for the convolution layers, while our proposed method uses the activation function $\operatorname{sign}(x)$ for all layers where activations are applied.


For biomedical image segmentation tasks, the proposed method outperforms the full-precision MC-Dropout method by up to $6.4$\% in terms of IoU score. In the worst scenario, our method results in a $69.69$\% reduction in the IoU score for the breast cancer dataset. Additionally, our approach outperforms the MC-Dropout method in most other metrics. Table~\ref{tab:segmentation} presents a summary of the results. 

The predictive performance is qualitatively shown in Figure~\ref{fig:Semantic_maps} for each dataset (with two examples). The sixth and third columns show the prediction mask for MC-Dropout and our method, respectively. It can be observed that the segmentation masks for the proposed method are similar to MC-Dropout and ground truth. In general, misclassified pixels are around the boundary of ground-truth masks. 

\subsubsection{Impact of MC Runs on Inference Accuracy}
\begin{figure}
    \centering
    \includegraphics[width=0.9\linewidth]{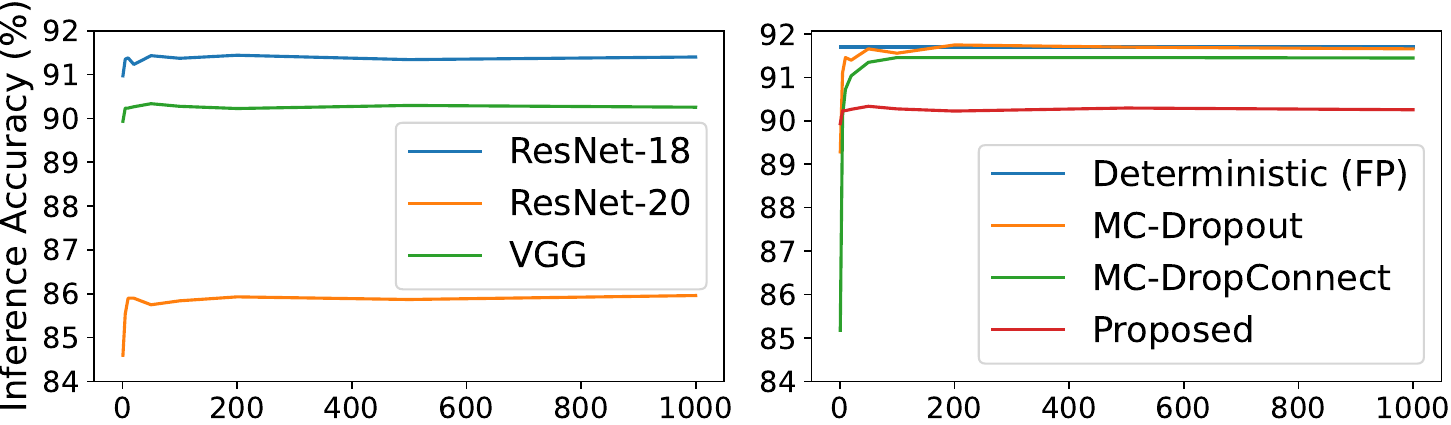}
    \caption{Evaluation of the effect of Monte Carlo runs on the inference accuracy of CIFAR-10 dataset on various topologies. }
    \label{fig:acc_vs_mc_runs}
\end{figure}

We observed that using Monte Carlo sampling ($T$ forward pass) for Bayesian inference generally enhances predictive performance across all datasets. For example, the inference accuracy of the ResNet-20 model increases from $84.63\%$ to $86.05\%$. In our evaluation, twenty samples (T = 20) for the larger model and fifty samples (T = 50) for the smaller model were used for Bayesian inference. { Fig.~\ref{fig:acc_vs_mc_runs} (a) shows that the proposed method requires a smaller number of samples, with inference accuracy plateaus around $T = 20$ to $50$. In comparison, MC-Dropout and MC-DropConnect methods require $100$, and $90$ Monte Carlo sampling, respectively, to achieve the maximum inference accuracy, as reported in~\cite{mobiny2021dropconnect} for CIFAR-10. In our experiment (see Fig.~\ref{fig:acc_vs_mc_runs} (b)), we observed that the MC-DropConnect method plateaus at $100$ Monte Carlo runs, and the MC-Dropout method plateaus at $200$ Monte Carlo runs on the same model (VGG) and dataset. Therefore, \emph{our method requires up to $180$ less Monte Carlo sampling}, leading to $10\times$ less XNOR and bit-counting operation, energy consumption, and latency for \emph{each Bayesian inference result}. We assume the same NN topology, hardware architecture, and memory device technology for a fair comparison. However, it should be noted that $T$ at which accuracy plateaus can vary from task to task and from model to model. }

\subsubsection{{Performance of BayBNN with Spin-ScaleDrop}}
We have shown that the predictive performance of our method is comparable to that of the full precision and binary implementations, assuming that the Spintronics Dropout module remains unchanged. However, it should also be tolerant to manufacturing and thermal variations in the Spintronic-based Dropout module.

\begin{table}[]
\setlength{\tabcolsep}{1pt}
\centering
\caption{{Evaluation of the inference accuracy of the proposed Spintronics-based Scale-Dropout with variations in the Dropout module. Variations in the probability of Dropout $p$ increased from $1\times$ to $3\times$ and the baseline model is the ideal scenario without variation.}
}
\centering
\resizebox{\linewidth}{!}{
\begin{tabular}{|c|c|ccc|ccc|}
\hline
\multirow{2}{*}{Topologie} & \multirow{2}{*}{Baseline} & \multicolumn{3}{c|}{Trained w/ Variations}                                                      & \multicolumn{3}{c|}{Trained w/o Variations}                                                     \\ \cline{3-8} 
                           &                                         & \multicolumn{1}{c|}{Var. $1\times$} & \multicolumn{1}{c|}{Var. $2\times$} & Var. $3\times$ & \multicolumn{1}{c|}{Var. $1\times$} & \multicolumn{1}{c|}{Var. $2\times$} & Var. $3\times$ \\ \hline
ResNet-18           &                  91.52\%           & \multicolumn{1}{c|}{91.78\%}        & \multicolumn{1}{c|}{91.77\%}        &    91.71\% & \multicolumn{1}{c|}{91.65\%}        & \multicolumn{1}{c|}{91.59\%}        &    91.58\%     \\ \hline
VGG                        &    90.45\%    & \multicolumn{1}{c|}{90.52\%}        & \multicolumn{1}{c|}{90.52\%}        &    90.55\%  & \multicolumn{1}{c|}{90.28\%}        & \multicolumn{1}{c|}{90.26\%}        &   90.30\%      \\ \hline
\end{tabular}
}
\label{tab:SpinscaleDrop_var_Acc}
\end{table}

To this end, we performed a small ablation study on the CIFAR-10 dataset with ResNet-18 and VGG topologies with models trained with and without variations in the Dropout module. Specifically, in one study, we trained both models with no variation in the probability of Dropout but, during Bayesian inference, we evaluated the model against our proposed Spintronic-based Dropout with up to $3\times$ the standard deviation $\sigma$ of the manufacturing variations. This means that the Dropout probability of each neuron can fluctuate by $ \pm10\%$ from the trained value. In this case, a slight improvement ($+ 0.13\%$) in predictive performance is observed for the ResNet-18 model, but for the VGG model, a slight reduction in inference accuracy ($- 0.19\%$) is observed. Nevertheless, the inference accuracy for both models remains close to the baseline accuracy. Furthermore, increasing the variation of a model from $0\times$ to $3\times$ has a negligible effect on the inference accuracy.

In the other case, the NN is trained considering the variation in the Dropout module (see Section~\ref{sec:spin_scale_drop_model}). 
In this case, unlike in the previous case, there is a slight improvement in predictive performance (up to $+ 0.26\%$) for both models compared to the baseline. Variation in the Dropout probability leads to more stochastically during Bayesian inference, and as a result, accuracy improves slightly. The results are summarized in Table~\ref{tab:SpinscaleDrop_var_Acc}.

\subsection{Uncertainty Estimation}\label{sec:uncer_eval}


\begin{figure}
\vspace*{-\baselineskip}

    \centering
    \includegraphics[width=\linewidth]{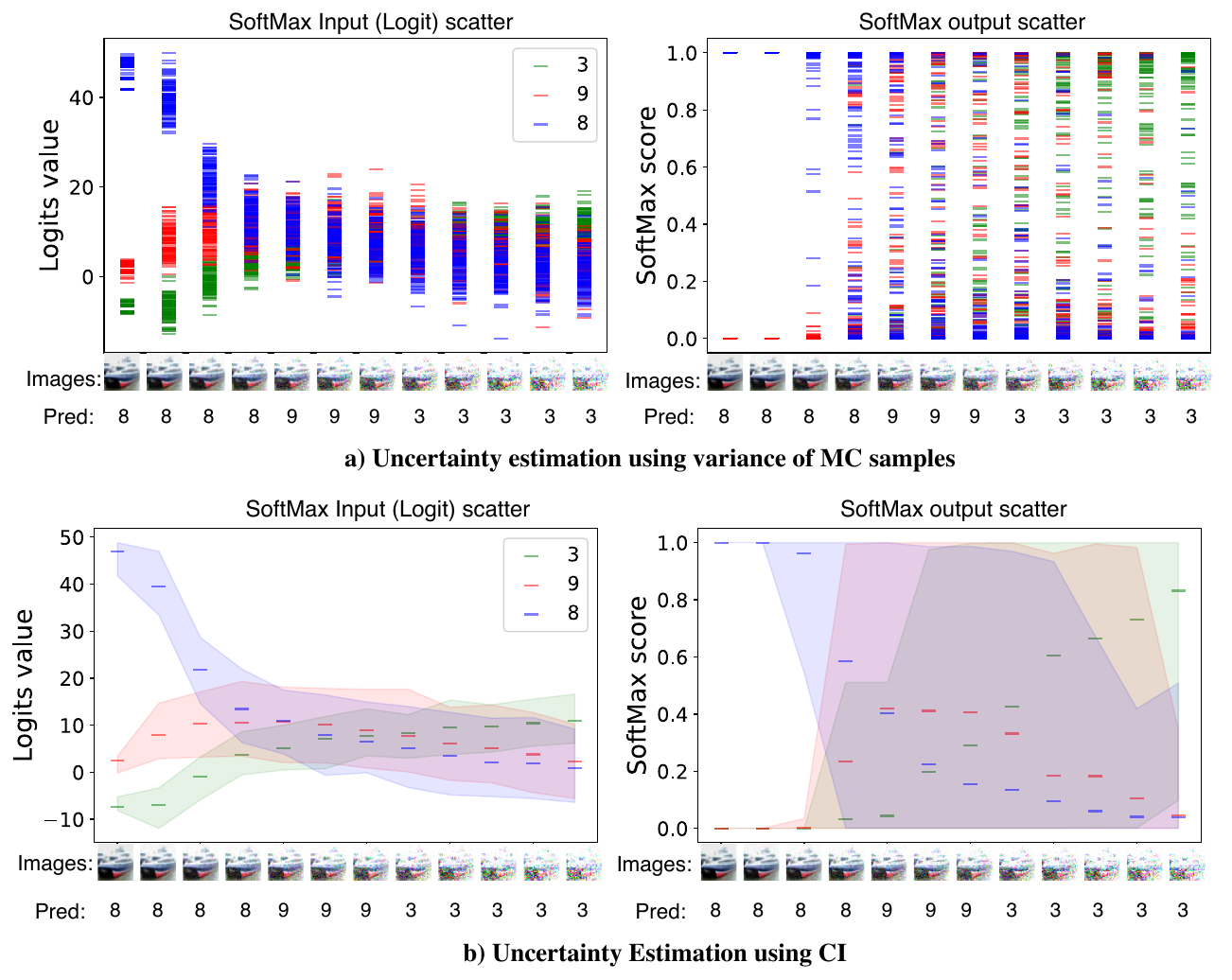}
    \caption{\footnotesize {\textbf{Detecting Distribution Shift on CIFAR-10}: a) A scatter and b) 95\% confidence interval of 100 forward passes of the softmax input (logits) and output for Scale-Dropout VGG topology. Uniform noise of increasing strength is added to a randomly sampled image of a ship (leveled as 8). The uncertainty of the prediction increases with the data distribution shift as shown by the high SoftMax scatter and the confidence interval. Although the model uncertainty is extremely high (best observed in color), the input for images 5 through 12 is classified as either a truck (leveled as 9) or a bird (leveled as 3).} }
    \label{fig:dist_shift_uniform_noise}
\end{figure}

\begin{figure}
\vspace*{-\baselineskip}
    \centering
    \footnotesize
    \includegraphics[width=\linewidth]{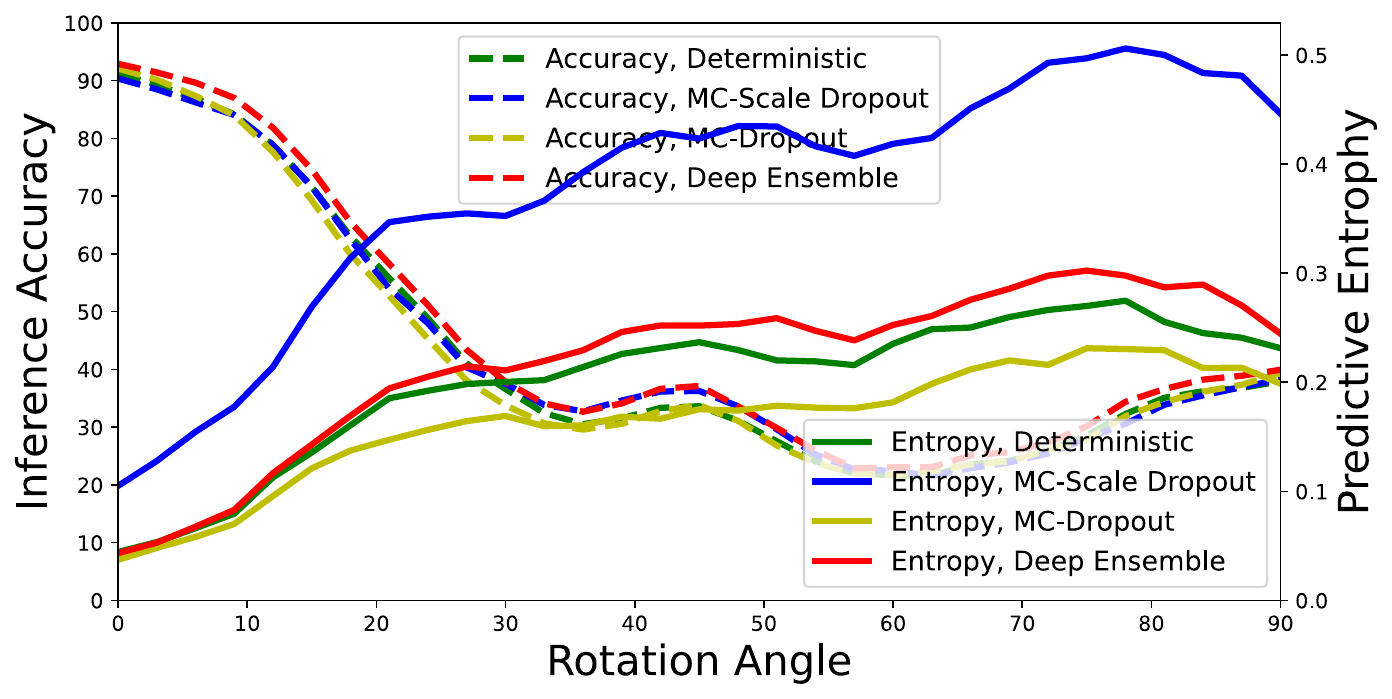}
    \caption{{The effect of distribution shift of inference images on inference accuracy (left y-axis) and predictive entropy (uncertainty estimate on right y-axis). Images are continuously rotated to introduce distribution shifts. The inference accuracy of all methods reduces with the same trend and the uncertainty of prediction increases with the data distribution shift. The uncertainty estimates of the proposed method outperform those of other methods, but the accuracy of the Ensemble method is higher in comparison.}
    }
    \label{fig:uncertainty_entrophy}
\end{figure}

\begin{table*}[]
\caption{The analysis of the proposed Scale-Dropout BayBNN on three biomedical segmentation datasets on SOTA topologies. A Dropout probability of $20\%$ is used with $20$ Monte Carlo samples for Bayesian inference. The best-performing matrices are in bold.
}
\resizebox{\linewidth}{!}{
\begin{tabular}{|c|c|c|c|c|c|c|c|c|c|c|}
\hline
Topology                         & Dataset                           & Method          & Bit-width (W/A) & Pixel-Acc        & IoU              & AUC              & F1               & Sensitivity      & Specificity      & Precision        \\ \hline
\multirow{2}{*}{U-Net}           & \multirow{3}{*}{Skin Cancer}            & FP      & 32/32           & 95.10\%          & 82.55\%          & 98.81\%          & 90.44\%          & 89.04\%          & 97.23\%          & 91.88\%          \\ \cline{3-11} 
                                 &                                  & MC-Dropout~\cite{gal2016Dropout}      & 32/32           & 95.05\%          & 81.67\%          & 98.86\%          & 89.91\%          & 84.74\%          & \textbf{98.67\%}         & \textbf{95.74\%}          \\ \cline{3-11} 
                                 &                                   & {Proposed} & 1/4             & \textbf{96.75\%} & \textbf{88.07\%} & \textbf{99.4\%} & \textbf{93.66\%} & \textbf{92.31\%} & {98.31\%} & {95.04\%} \\ \hline
\multirow{3}{*}{Bayesian SegNet} & \multirow{4}{*}{Breast Cancer}     & FP      & 32/32           & 97.47\%          & 67.65\%          & 96.60\%          & 80.71\%          & 76.95\%          & 98.99\%          & 84.84\%          \\ \cline{3-11} 
                                 &                                   & MC-Dropout~\cite{gal2016Dropout}      & 32/32           & \textbf{97.83\%}          & \textbf{71.82\%}          & 96.40\%          & \textbf{83.21\%}          & \textbf{78.23\%}          & \textbf{99.28\%}          & \textbf{88.86\%}          \\ \cline{3-11} 
                                 &                                   & Spatial-SpinDrop~\cite{soyed_TNANO23}  & 1/4             & {96.32\%} & {56.12\%} & {94.31\%} & {71.90\%} & {68.50\%} & {98.37\%} & {75.64\%} \\  \cline{3-11} 
                                 &                                   & SpinDrop~\cite{soyed_nanoarch22, ahmed2023spindrop}  & 1/4             & {97.47\%} & {68.17\%} & {95.48\%} & {80.13\%} & {74.25\%} & {99.18\%} & {87.02\%} \\  \cline{3-11} 
                                 &                                   & {Proposed}  & 1/4             & {97.69\%} & {69.69\%} & \textbf{96.69\%} & \textbf{82.14\%} & {77.19\%} & {99.21\%} & {87.76\%} \\ 
                                 \hline
\multirow{2}{*}{Bayesian SegNet} & \multirow{2}{*}{COVID-19 Lung CT} & FP      & 32/32           & 99.53\%          & 72.86\%          & 99.11\%          & 84.30\%          & 83.17\%          & \textbf{99.78\%}          & \textbf{85.46\%}          \\ \cline{3-11} 
                                 &                                   & MC-Dropout~\cite{gal2016Dropout}      & 32/32           & 99.51\%          & 72.88\%          & 99.39\%          & 84.32\%          & \textbf{86.99\%}          & 99.7\%          & 81.8\%          \\ \cline{3-11} 
                                 &                                   & {Proposed} & 1/4             & \textbf{99.55\%} & \textbf{74.43\%} & \textbf{99.69\%} & \textbf{85.34\%} & {85.84\%} & {99.76\%} & {84.85\%} \\ \hline
\end{tabular}
}
\label{tab:segmentation}
\end{table*}

\begin{figure}
    \centering
    \includegraphics[width=\linewidth]{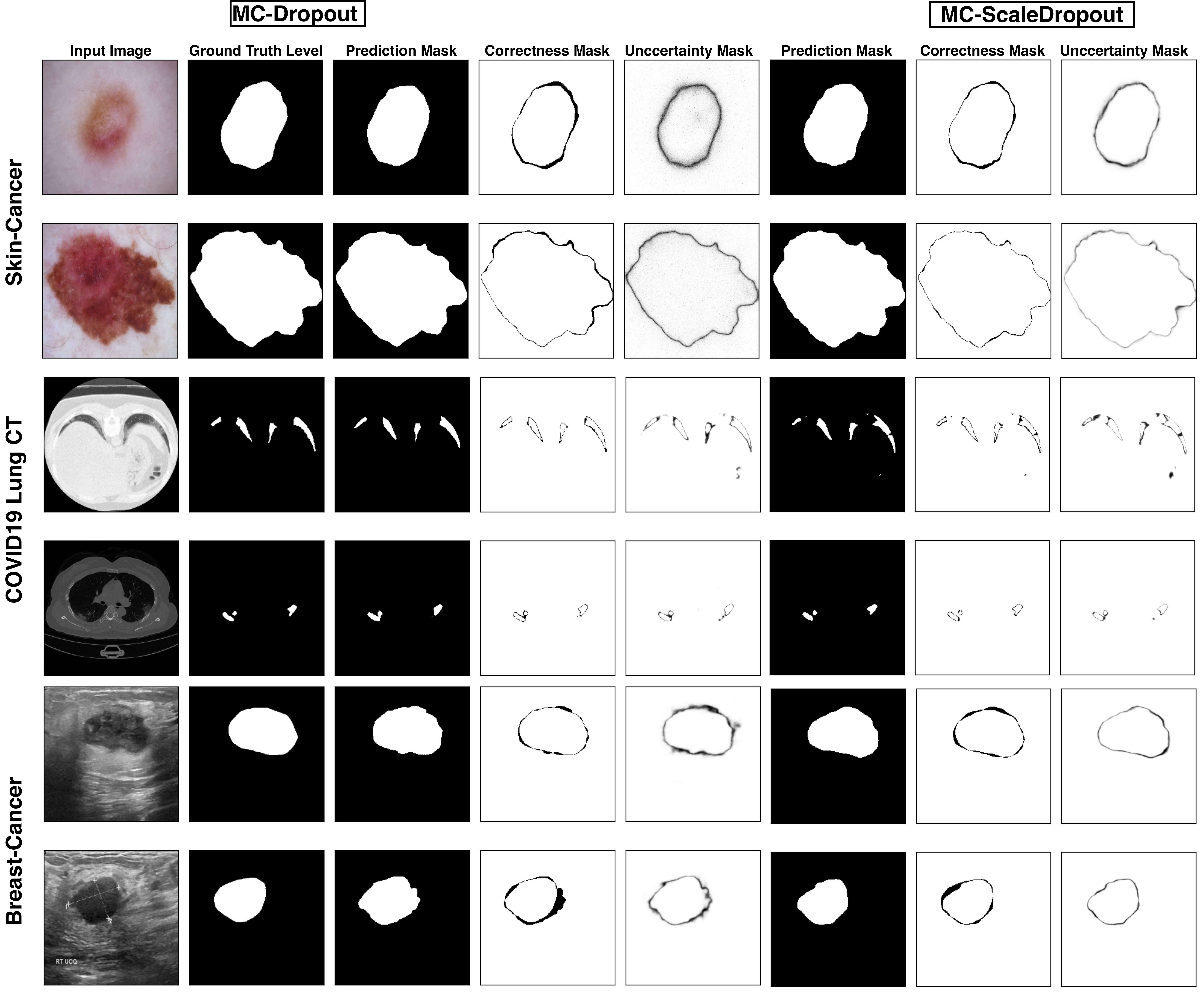}
    \caption{\footnotesize The outcomes of semantic segmentation and uncertainty estimations for the Skin-Cancer, COVID-19 Lung-CT, and Breast Cancer datasets. Each row comprises the input image followed by the ground truth, prediction mask, correctness mask, and uncertainty mask for both the MC-Dropout and proposed MC-Scale Dropout methods. The correctness mask is a binary representation of accurate and inaccurate predictions. The uncertainty mask is the normalized [$0$, $1$] uncertainty mask derived from twenty Monte Carlo samples. For prediction masks, a $0.5$ threshold is used. On the correctness and uncertainty masks, the correct and certain regions are depicted in white. Similarly, a region that is incorrect or uncertain is displayed in black.}
    \label{fig:Semantic_maps}
\end{figure}

\paragraph{Detecting Distribution Shift}
To show the effectiveness of the proposed Scale-Dropout method in detecting distribution shifts in the data, we conducted two experiments. In one experiment, we continuously added random noise from a uniform distribution to the input data with increasing strength. As shown in Fig.~\ref{fig:dist_shift_uniform_noise}, the variance and confidence interval in the model logits (SoftMax input) and the predicted probability of the output classes (SoftMax output) increases as the strength of noise increases. In other words, the uncertainty in the prediction increases as the distribution dataset shifts away from the original distribution. However, despite the high uncertainty, the model predicts a truck or a bird.

On the other hand, we have performed another experiment with all images of the CIFAR-10 dataset on the VGG model continuously rotated up to $90^\circ$. It can be seen in Fig.~\ref{fig:uncertainty_entrophy} that as the images are rotated, the inference accuracy decreases and the predictive entropy increases from the starting entropy. Our method is compared with deterministic as well as common uncertainty estimation techniques, namely MC-Dropout~\cite{gal2016Dropout} and Deep Ensemble~\cite{lakshminarayanan2017simple} with five randomly initialized models. The trend of decrease in inference accuracy is similar for all models. {However, the Deep Ensemble slightly outperforms the proposed and other methods in terms of accuracy. Regardless, our proposed MC Scale-Dropout method produces significantly more predictive entropy compared to other methods, including the Deep Ensemble method. This is because the proposed MC Scale-Dropout method effectively turns a single model into numerous ensembles by enabling Dropout during inference, allowing it to generate multiple predictions from a single model. Whereas, Deep Ensemble has limited models in the ensemble, e.g., five models. Also, the scale Dropout provides a regularization effect that can sometimes lead to better generalization in MC Scale-Dropout models compared to individual models in a Deep Ensemble.} Consequently, our method can produce better uncertainty estimates compared to related works even with a 1-bit model.

Note that in the MC-Dropout model, Dropout is applied to the extracted features from the convolutional layers to achieve similar inference accuracy. If Dropout is applied to all layers, the predictive entropy increases, but inference accuracy decreases significantly, e.g., by more than 3\%. Our approach provides a good balance between uncertainty estimates without any degradation in accuracy.

\begin{table}
\setlength{\tabcolsep}{1pt}
\footnotesize
\caption{Evaluation of the proposed MC-Scale Dropout method in detecting OOD across various topologies. All the models are trained on CIFAR-10 dataset.}
\resizebox{\linewidth}{!}{
\begin{tabular}{|c|c|c|c|c|c|c|c|}
\hline
Topology   & Method  & $\hat{\mathcal{D}}_1$     & $\hat{\mathcal{D}}_2$   & $\hat{\mathcal{D}}_3$     & $\hat{\mathcal{D}}_4$     & $\hat{\mathcal{D}}_5$     & $\hat{\mathcal{D}}_6$     \\ \hline
\hline
\multirow{3}{*}{VGG} & \textbf{Proposed} & ${95.36\%}$ & ${95.40\%}$ & $\mathbf{96.19\%}$ & $\mathbf{88.73\%}$ & $\mathbf{98.02\%}$ & ${85.24\%}$ \\
& Spatial-SpinDrop~\cite{soyed_TNANO23} & $\mathbf{99.99\%}$ & $\mathbf{100\%}$ & $92.9\%$  & $78.91\%$ & $99.81\%$ & $\mathbf{100\%}$ \\
& SpinBayes~\cite{ahmed_spinbayes_2023} & $99.86\%$ & $-$ & $94.35\%$  & $ -$ & $80.33\%$ & $89.31\%$ \\
\hline
\multirow{2}{*}{ResNet-18} & \textbf{Proposed} & $\mathbf{100\%}$ & $\mathbf{100\%}$ & ${97.49\%}$ & ${77.77\%}$ & ${91.53\%}$ & ${78.61\%}$ \\
& Spatial-SpinDrop~\cite{soyed_TNANO23} & $100\%$   & $100\%$ & $\mathbf{100\%}$   & $\mathbf{92.26\%}$   & $\mathbf{99.98\%}$   & $\mathbf{97.39\%}$ \\
\hline
\multirow{2}{*}{ResNet-20} & \textbf{Proposed} & ${96.51\%}$ & $\mathbf{100\%}$ & ${90.34\%}$ & $\mathbf{93.55\%}$ & $\mathbf{100\%}$ & ${99.8\%}$ \\
& Spatial-SpinDrop~\cite{soyed_TNANO23}& $\mathbf{97.2\%}$  & $100\%$ & $\mathbf{90.79\%}$ & $87.94\%$ & $99.03\%$ & $\mathbf{99.81\%}$ \\
\hline
\end{tabular}
}
\label{tab:ood}
\end{table}

\paragraph{Detecting Out-of-distribution Data}

We show that the model uncertainty increases as the distribution of the data shifts from the original distribution. Here, we perform an ablation study with six (definitive) out-of-distribution datasets.

As depicted in Table~\ref{tab:ood}, our proposed method can achieve a detection rate of OOD of up to $100\%$ across various model architectures and six different OOD datasets ($\hat{\mathcal{D}}_1$ through $\hat{\mathcal{D}}_6$). There are some variations in OOD detection rates across different architectures for the same OOD dataset. However, even in these cases, our method can consistently achieve a high OOD detection rate, with the lowest detection rate being $77.77\%$ on the ResNet-18 model with $\hat{\mathcal{D}}_4$ dataset. However, when the threshold for SoftMax confidence increases from $95\%$ to $99\%$, the OOD detection rate in the dataset $\hat{\mathcal{D}}_4$ improved to $81.78\%$, an $\sim 4\%$ improvement. Compared to MC-Spatial Dropout and SpinBayes methods, the OOD detection rates are generally similar. In the worst case, the OOD detection rate is $\sim 14\%$ lower for the VGG topology on the $\hat{\mathcal{D}}_6$ dataset.
Therefore, the results indicate that the proposed MC-Scale Dropout method is a robust and reliable solution to OOD detection across diverse model architectures and datasets.

\paragraph{Epistemic Uncertainty of Semantic Segmentaion}

For biomedical segmentation tasks, the epistemic uncertainty is calculated for each pixel. The fifth and eighth columns of Fig.~\ref{fig:Semantic_maps} depict the pixel-wise uncertainty masks (qualitatively) for the MC-Dropout and the proposed MC-Scale Dropout method. In segmentation tasks, an ideal model would produce high uncertainty around misclassified pixels and low uncertainty around correctly classified pixels. Overall, as depicted in Fig.~\ref{fig:Semantic_maps}, the uncertainty is high around the misclassified pixels for the proposed method, but correctly classified pixels have low uncertainty. In general, the uncertainty masks for MC-Dropout are darker, depicting slightly stronger uncertainty estimates due to their higher model precision (32 bits) and a higher Dropout probability (50\%). However, in some cases, the uncertainty mask is also stronger in the region of correctly classified pixels. However, our proposed method produces uncertainty only around miss-classified pixels.

\subsection{Hardware Overhead Analysis}
To assess the energy consumption of the proposed approach, we estimated the required resources for implementing a network of five layers with the Scale-Dropout method, and we assumed using $10$ crossbar arrays of $256\times256$ and 10 Spin-ScaleDrop modules to implement a LeNet-5 network. The total area needed for the implementation of the LeNet-5 topology is $0.401 \, \text{mm}^2$ comprising the crossbar arrays and the memories. The area estimation is based on the NVSim and layout measurement. 
Given the energy consumption of the different components of our architecture shown in Table~\ref{tab:circuit_data}. We used the NVSim simulator to estimate the total energy consumption for an inference run and multiplied this value by the number of forward passes (MC run). The analysis is carried out for ten forward passes ($T=10$). The energy consumption of an inference run is shown in Table~\ref{tab:energy_sota} compared to other FPGA and CIM implementations. 
{We evaluated two topologies LeNet-5 for the MNIST dataset and VGG small for CIFAR-10. For a consistent benchmark, the same metrics as in previous studies were used.}
The Scale-Dropout approach significantly improves energy efficiency, reaching up to $100\times$ higher efficiency compared to the method presented in \cite{awano_bynqnet_2020}. Compared to the implementation in~\cite{malhotra_exploiting_2020}, our approach is $51\times$ better.  Furthermore, compared to the implementation based on STT-MRAM~\cite{soyed_TNANO23}, the proposed approach exhibits $3.77\times$ better efficiency. Finally, compared to reference~\cite{yang_all-spin_2020}, our approach demonstrates $4.38\times$ greater energy efficiency. {To scale up the approach, we have performed an energy consumption estimation with a VGG small topology and we report  0.29 $\mu J/Image$. Thus, energy consumption remains notably low even when considering a larger dataset such as CIFAR-10. }

Furthermore, Scale-Dropout requires only one RNG per layer compared to similar approaches~\cite{soyed_nanoarch22,soyed_TNANO23}. An RNG can be shared for all layers to reduce the number of RNGs for the whole model to one. This significantly contributes to a reduction in energy consumption. 
\begin{table}{}
\centering
\caption{Layer-wise overhead analysis of the proposed method in comparison to existing works SpinDrop~\cite{soyed_nanoarch22} and Spatial-SpinDrop~\cite{soyed_TNANO23}.}
\resizebox{\linewidth}{!}{
\begin{tabular}{|cccccc|}
\hline
\multicolumn{6}{|c|}{Layer-wise application of scale Dropout}                                                                                                                                                                                                                                                                                                                                                                                    \\ \hline
\multicolumn{1}{|c|}{Method}                             & \multicolumn{1}{c|}{\begin{tabular}[c]{@{}c@{}}Mapping \\ Strategy\end{tabular}}   & \multicolumn{1}{c|}{\begin{tabular}[c]{@{}c@{}}\# of Dropout \\ Modules\end{tabular}} & \multicolumn{1}{c|}{Area} & \multicolumn{1}{c|}{\begin{tabular}[c]{@{}c@{}}Power\\ Consumption\end{tabular}} & \begin{tabular}[c]{@{}c@{}}Sampling \\ Latency\end{tabular} \\ \hline
\multicolumn{1}{|c|}{\multirow{2}{*}{SpinDrop~\cite{soyed_nanoarch22}}}          & \multicolumn{1}{c|}{\circled{1}}                                                             & \multicolumn{1}{c|}{$K*K*C_{in}$}                                                     & \multicolumn{1}{c|}{$79833.6\mu m^2$}     & \multicolumn{1}{c|}{$51.84 mW$}                                                             & $15 ns$                                                            \\ \cline{2-6} 
\multicolumn{1}{|c|}{}                                   & \multicolumn{1}{c|}{\circled{2}}                                                             & \multicolumn{1}{c|}{$K*K*C_{in}$}                                                     & \multicolumn{1}{c|}{$79833.6\mu m^2$}     & \multicolumn{1}{c|}{$51.84 mW$}                                                             &   $15 ns$                                                           \\ \hline
\multicolumn{1}{|c|}{\multirow{2}{*}{Spatial-SpinDrop~\cite{soyed_TNANO23}}} & \multicolumn{1}{c|}{\circled{1}}                                                             & \multicolumn{1}{c|}{$C_{in}$}                                                         & \multicolumn{1}{c|}{$8870.4 \mu m^2$}     & \multicolumn{1}{c|}{$5.76 mW$}                                                             &       $15 ns$                                                       \\ \cline{2-6} 
\multicolumn{1}{|c|}{}                                   & \multicolumn{1}{c|}{\circled{2}}                                                             & \multicolumn{1}{c|}{$C_{in}$}                                                         & \multicolumn{1}{c|}{$8870.4 \mu m^2$}     & \multicolumn{1}{c|}{$5.76 mW$}                                                             &       $15 ns$                                                      \\ \hline
\multicolumn{1}{|c|}{\multirow{2}{*}{\textbf{Proposed}}} & \multicolumn{1}{c|}{\circled{1}}                                                             & \multicolumn{1}{c|}{$\textbf{1}$}                                                         & \multicolumn{1}{c|}{$\mathbf{34.65\mu m^2}$}     & \multicolumn{1}{c|}{$\textbf{0.0225 mW}$}                                                             &       $15 ns$                                                       \\ \cline{2-6} 
\multicolumn{1}{|c|}{}                                   & \multicolumn{1}{c|}{\circled{2}}                                                             & \multicolumn{1}{c|}{$\textbf{1}$}                                                         & \multicolumn{1}{c|}{$\mathbf{34.65\mu m^2}$}     & \multicolumn{1}{c|}{$\textbf{0.0225 mW}$}                                                             &       $15 ns$                                                      \\ \hline
\multicolumn{6}{|c|}{Topology-wise application of scale Dropout}                                                                                                                                                                                                                                                                                                                                                                                 \\ \hline
\multicolumn{1}{|c|}{Method}                             & \multicolumn{1}{c|}{\begin{tabular}[c]{@{}c@{}}Adaptive \\ Avg. Pool\end{tabular}} & \multicolumn{1}{c|}{\begin{tabular}[c]{@{}c@{}}\# of Dropout\\ Modules\end{tabular}}  & \multicolumn{1}{c|}{Area} & \multicolumn{1}{c|}{\begin{tabular}[c]{@{}c@{}}Power\\ Consumption\end{tabular}}  & \begin{tabular}[c]{@{}c@{}}Sampling\\ Latency\end{tabular}  \\ \hline
\multicolumn{1}{|c|}{\multirow{2}{*}{SpinDrop~\cite{soyed_nanoarch22}}}          & \multicolumn{1}{c|}{Used}                                                          & \multicolumn{1}{c|}{$C_{out}$}                                                         & \multicolumn{1}{c|}{$17740.8 \mu m^2$}     & \multicolumn{1}{c|}{$11.52 mW$}                                                      &    $15 ns$                                                        \\ \cline{2-6} 
\multicolumn{1}{|c|}{}                                   & \multicolumn{1}{c|}{Not Used}                                                      & \multicolumn{1}{c|}{$K*K*C_{out}$}                                                     & \multicolumn{1}{c|}{$159 667.2\mu m^2$}     & \multicolumn{1}{c|}{$103.68 mW$}                                                             &                                        $15 ns$                      \\ \hline
\multicolumn{1}{|c|}{\multirow{2}{*}{Spatial-SpinDrop~\cite{soyed_TNANO23}}}          & \multicolumn{1}{c|}{Used}                                                          & \multicolumn{1}{c|}{$C_{out}$}                                                         & \multicolumn{1}{c|}{$17740.8 \mu m^2$}     & \multicolumn{1}{c|}{$11.52 mW$}                                                             & $15 ns$                                                            \\ \cline{2-6} 
\multicolumn{1}{|c|}{}                                   & \multicolumn{1}{c|}{Not Used}                                                      & \multicolumn{1}{c|}{$C_{out}$}                                                     & \multicolumn{1}{c|}{$17740.8 \mu m^2$}     & \multicolumn{1}{c|}{$11.52 mW$}                                                             &                                                       $15 ns$       \\ \hline
\multicolumn{1}{|c|}{\multirow{2}{*}{\textbf{Proposed}}}          & \multicolumn{1}{c|}{Used}                                                          & \multicolumn{1}{c|}{$\textbf{1}$}                                                         & \multicolumn{1}{c|}{$\mathbf{34.65\mu m^2} $}     & \multicolumn{1}{c|}{$\textbf{0.0225 mW}$}                                                             & $15 ns$                                                            \\ \cline{2-6} 
\multicolumn{1}{|c|}{}                                   & \multicolumn{1}{c|}{Not Used}                                                      & \multicolumn{1}{c|}{$\textbf{1}$}                                                     & \multicolumn{1}{c|}{$\mathbf{34.65\mu m^2} $}     & \multicolumn{1}{c|}{$\textbf{0.0225 mW}$}                                                             &                                                       $15 ns$       \\ \hline
\end{tabular}}

\label{tab:drop_modules}
\end{table}

\begin{table}[hbt]
\caption{Energy Efficiency Comparison of Hardware Implementations}
\resizebox{\columnwidth}{!}
{
\centering
\begin{tabular}{|l|l|l|l|l|}
\hline
Related works     & Technology & {Topology} & Bit resolution & Energy \\ \hline
{H. Fan et al.\cite{fan_fpga-based_2022}} & {FPGA}   & {ResNet18}& {8-bit} &   {0.014~\SI{}{\joule}/Image}  \\ \hline
R.Cai et al.\cite{cai_vibnn_2018}      & FPGA& {3-FC} & 8-bit         & 18.97~\SI{}{\micro\joule}/Image         \\ \hline
X.Jia et al.\cite{jia_efficient_2021}      & FPGA& {3-FC} & 8-bit        & 46.00~\SI{}{\micro\joule}/Image  \\ \hline
H.Awano et al. \cite{awano_bynqnet_2020}    & FPGA&  {3-FC}    & 7-bit     & 21.09~\SI{}{\micro\joule}/Image           \\ \hline
A. Malhotra et al. \cite{malhotra_exploiting_2020}      & RRAM& {3-FC} & 4-bit        & 9.30~\SI{}{\micro\joule}/Image         \\ \hline
S.T.Ahmed et al.\cite{soyed_nanoarch22} & STT-MRAM&  {LeNet-5}& 1-bit     & 2.00~\SI{}{\micro\joule}/Image              \\ \hline
S.T.Ahmed et al.\cite{soyed_TNANO23} & STT-MRAM&  {LeNet-5} & 1-bit     & 0.68~\SI{}{\micro\joule}/Image  \\ \hline
K.Yang et al.\cite{yang_all-spin_2020} & Domain wall-MTJ & {3-FC}  & {4-bit}     & {0.79}~\SI{}{\micro\joule}/Image  \\ \hline
\textbf{Proposed implementation (MNIST)} & \textbf{SOT-MRAM} & {\textbf{LeNet-5}}  & \textbf{1-bit} & \textbf{0.18}~\SI{}{\micro\joule}/Image  \\ \hline
{\textbf{Proposed implementation (CIFAR-10)}} & {\textbf{SOT-MRAM}} & {\textbf{VGG}}  & {\textbf{1-bit}} & {\textbf{0.29}~\SI{}{\micro\joule}/Image}  \\ \hline

\end{tabular}
}
\label{tab:energy_sota}
\end{table}

\section{Discussion}\label{sec:discussion}
\subsection{In Distribution Uncertainty Analysis}
{We thoroughly analyzed the performance of the proposed method in data distribution shift and out-of-distribution data in Section~\ref{sec:uncer_eval}, it is equally important to perform well when it receives in-distribution data. This means that correct predictions should have low uncertainty and a model should accept most of them.}

{In our in-distribution data analysis (Table \ref{tab:ID_analysis}), we present the accepted, rejected, TPR, TNR, and AR percentages. TPR indicates the rate of correct and accepted predictions, while TNR refers to rejected and incorrect predictions. High TPR and TNR rates are desired as they suggest that most of the accepted predictions have low uncertainty, and incorrect predictions have high uncertainty. A high AR rate also indicates that most of the correct predictions are accepted.}

{The VGG and ResNet-18 models, with their larger size, effectively handle the complexity of the CIFAR-10 task, showing acceptance of approximately 80\% and more than 80\% in both TPR and TNR, plus more than 97\% in AR, confirming the efficacy of our method.}

{On the contrary, the smaller ResNet-20 model is not optimal for handling the complexity of CIFAR-10, leading to 'uncertainty in model architecture' \cite{he2023survey} and consequently to greater uncertainty in prediction. To be specific, its inference accuracy is comparatively lower $\sim 86\%$ compared to $\sim 91\%$ for the other model, since it only has 16, 32, and 64 neurons in the residual blocks. Thus, it has a lower acceptance rate (41\%). That means that most predictions are uncertain and our method is also effective in quantifying 'uncertainty in the model architecture'. }

{Note that our classification of the predictions (OOD or ID) with our approach (see Equation~\ref{eq}) is conservative and prioritizes certainty. Adjusting quantile and confidence scores (see Equation 24) can increase acceptance rates closer to inference accuracy but may decrease OOD detection rates.}

\begin{table}[]
\centering
\caption{{Analysis of the proposed method using in-distribution data, showing True Positive Rate (TPR), True Negative Rate (TNR), and Acceptance Rate (AR) for various topologies.}
}
\resizebox{\linewidth}{!}{
{
\begin{tabular}{|c|c|c|c|c|c|}
\hline
Topology  & Accept & Reject & TPR   & TNR   & AR    \\ \hline
VGG       & 77.43\%  & 22.57\%  & 84.29\% & 81.67\% & 97.45\% \\ \hline
ResNet-18 & 77.48\%  & 22.52\%  & 84.00\% & 81.83\% & 97.67\% \\ \hline
ResNet-20 & 41.00\%   & 58.98\%  & 59.24\% & 95.83\% & 99.12\% \\ \hline
\end{tabular}
}
}
\label{tab:ID_analysis}
\end{table}

\subsection{Corruption Robustness Analysis}


{The proposed method is evaluated on 15 common corruptions reported in the work (CIFAR-10-C)~\cite{hendrycks2019benchmarking} with various topologies with and without pre-processing, as shown in Fig.~\ref{tab:corr_robust}. Our approach can achieve an OOD detection rate of on average $87.06\%$, $86.10\%$ and $97.64\%$ for VGG, ResNet-18, and ResNet-20 topologies, respectively, when no pre-processing is applied.}

{On the other hand, when the corruption robustness dataset is pre-processed by channel-wise normalizing them, i.e., they have the same channel-wise distribution as the clean CIFAR-10 data the model expects, the corruption error drastically reduces. For example, the mean corruption error for VGG was reduced from $82.84\%$ to $49.95\%$. Consequently, the uncertainty of the predictions also reduces. Specifically, our approach achieves OOD detection rates of $58.48\%$, $56.21\%$, and $87.73\%$, respectively, for VGG, ResNet-18 and ResNet-20 topologies.
Therefore, pre-processing the dataset standardizes the data and improves the corruption robustness similar to the histogram equalization results reported in the work}~\cite{hendrycks2019benchmarking}.

{In terms of topology, in larger networks, e.g., ResNet-18, the corruption error is relatively lower. For example, in the case of Gaussian noise, the corruption error is reduced from 86.85\% in VGG to 83.52\% in ResNet-18. A similar trend is observed for other datasets. However, despite the fact that the ResNet-20 model is smaller than ResNet-18, it has a relatively higher corruption error because the smaller model introduces “uncertainty in model architecture” as mentioned in the previous section. }

{Nevertheless, there is a direct relationship between corruption error, uncertainty, and, in turn, the OOD detection rate.  In cases where the accuracy is reduced by a small margin, the model uncertainty is low, and the OOD detection rate with our approach is also low. For example, the worst-case OOD detection rate for the VGG topology is $29.53\%$. This is achieved when the accuracy is reduced by only $6.89\%$ for brightness corruption. On the other hand, the highest OOD detection rate is achieved when the accuracy is reduced by $78.88\%$ for the VGG topologies.
}

\subsection{Variability and scalability}

{This study extends previous research demonstrating the robustness of the dropout approach against device variability \cite{ahmed2023spindrop}.It highlights the impact of dropout module variations on network accuracy as shown in Table~\ref{tab:SpinscaleDrop_var_Acc}. Moreover, we propose to utilize SOT-MRAM, capable of achieving resistance levels up to several M$\Omega$, which aligns with previous simulations emphasizing resistance's crucial role in constructing large arrays~\cite{ielmini_status_2022}, validating the scalability and energy efficiency of this approach.}

\begin{table*}[]
\setlength{\tabcolsep}{1pt}
\footnotesize
\caption{{Analysis of mean corruption errors (mCE) and mean out-of-distribution detection (mOOD) detection values of different topologies when various corruptions applied CIFAR-10 with and without pre-processing (PP). All numbers represent percentages.}
}
\centering
\resizebox{\linewidth}{!}{
{
\begin{tabular}{ccccc|ccc|cccc|cccc|cccc|}
                                                                                            & \multicolumn{1}{l}{}     &                                             &                            &       & \multicolumn{3}{c|}{Noise}                                                              & \multicolumn{4}{c|}{Blur}                                                                                                                         & \multicolumn{4}{c|}{Weather}                                                                                                                      & \multicolumn{4}{c|}{Digital}                                                                                                                      \\ \hline
\multicolumn{1}{|c|}{Topology}                                                              & \multicolumn{1}{c|}{PP}  & \multicolumn{1}{c|}{Error}                  & \multicolumn{1}{c|}{mCE}   & mOOD  & \multicolumn{1}{c|}{Gauss.}        & \multicolumn{1}{c|}{Shot}          & Impulse       & \multicolumn{1}{c|}{Defocus}       & \multicolumn{1}{c|}{Glass}         & \multicolumn{1}{c|}{Motion}        & Zoom                               & \multicolumn{1}{c|}{Snow}          & \multicolumn{1}{c|}{Frost}         & \multicolumn{1}{c|}{Fog}           & Bright                             & \multicolumn{1}{c|}{Contrast}      & \multicolumn{1}{c|}{Elastic}       & \multicolumn{1}{c|}{Pixel}         & JPEG                               \\ \hline
\multicolumn{1}{|c|}{\multirow{2}{*}{VGG}}                                                  & \multicolumn{1}{c|}{No}  & \multicolumn{1}{c|}{\multirow{2}{*}{9.55}}  & \multicolumn{1}{c|}{82.84} & 87.06 & \multicolumn{1}{c|}{80.57 (86.85)} & \multicolumn{1}{c|}{80.08 (86.27)} & 71.6 (61.07)  & \multicolumn{1}{c|}{91.41 (86.91)} & \multicolumn{1}{c|}{89.99 (86.41)} & \multicolumn{1}{c|}{88.95 (86.03)} & 95.24 (86.45)                      & \multicolumn{1}{c|}{76.38 (82.33)} & \multicolumn{1}{c|}{87.52 (85.51)} & \multicolumn{1}{c|}{97.42 (88.43)} & 86.23 (70.71)                      & \multicolumn{1}{c|}{84.98 (90.16)} & \multicolumn{1}{c|}{95.26 (82.59)} & \multicolumn{1}{c|}{87.47 (81.12)} & 92.85 (81.83)                      \\ \cline{2-2} \cline{4-20} 
\multicolumn{1}{|c|}{}                                                                      & \multicolumn{1}{c|}{Yes} & \multicolumn{1}{c|}{}                       & \multicolumn{1}{c|}{49.95} & 58.48 & \multicolumn{1}{c|}{68.39 (70.45)} & \multicolumn{1}{c|}{67.96 (66.54)} & 78.78 (87.85) & \multicolumn{1}{c|}{62.75 (50.72)} & \multicolumn{1}{c|}{59.96 (50.8)}  & \multicolumn{1}{c|}{55.78 (43.25)} & 58.97 (46.06)                      & \multicolumn{1}{c|}{47.0 (34.86)}  & \multicolumn{1}{c|}{56.7 (46.96)}  & \multicolumn{1}{c|}{58.09 (41.78)} & 29.53 (16.44)                      & \multicolumn{1}{c|}{89.84 (78.31)} & \multicolumn{1}{c|}{48.13 (30.07)} & \multicolumn{1}{c|}{47.38 (55.28)} & 47.95 (29.93)                      \\ \hline
\multicolumn{1}{|c|}{\multirow{2}{*}{\begin{tabular}[c]{@{}c@{}}ResNet\\ -18\end{tabular}}} & \multicolumn{1}{c|}{No}  & \multicolumn{1}{c|}{\multirow{2}{*}{8.48}}  & \multicolumn{1}{c|}{79.49} & 86.10 & \multicolumn{1}{c|}{84.38 (83.52)} & \multicolumn{1}{c|}{83.57 (82.03)} & 82.66 (84.26) & \multicolumn{1}{l|}{95.89 (82.13)} & \multicolumn{1}{l|}{89.60 (83.06)} & \multicolumn{1}{l|}{93.37 (79.46)} & \multicolumn{1}{l|}{94.48 (81.61)} & \multicolumn{1}{l|}{80.01 (72.81)} & \multicolumn{1}{l|}{89.97 (80.81)} & \multicolumn{1}{l|}{98.64 (82.34)} & \multicolumn{1}{l|}{73.90 (61.02)} & \multicolumn{1}{c|}{82.66 (84.26)} & \multicolumn{1}{l|}{86.98 (78.20)} & \multicolumn{1}{l|}{74.18 (81.28)} & \multicolumn{1}{l|}{81.25 (75.56)} \\ \cline{2-2} \cline{4-20} 
\multicolumn{1}{|c|}{}                                                                      & \multicolumn{1}{c|}{Yes} & \multicolumn{1}{c|}{}                       & \multicolumn{1}{c|}{47.32} & 56.21 & \multicolumn{1}{c|}{66.53 (66.71)} & \multicolumn{1}{c|}{60.42 (65.96)} & 73.73 (59.13) & \multicolumn{1}{c|}{62.75 (50.72)} & \multicolumn{1}{l|}{58.90 (49.82)} & \multicolumn{1}{l|}{59.54 (44.34)} & \multicolumn{1}{l|}{61.11 (47.05)} & \multicolumn{1}{l|}{47.7 (32.58)}  & \multicolumn{1}{l|}{53.78 (59.25)} & \multicolumn{1}{l|}{58.09 (41.78)} & \multicolumn{1}{l|}{28.79 (14.61)} & \multicolumn{1}{l|}{65.71 (74.63)} & \multicolumn{1}{l|}{47.98 (26.66)} & \multicolumn{1}{l|}{55.15 (53.48)} & \multicolumn{1}{l|}{42.98 (23.02)} \\ \hline
\multicolumn{1}{|c|}{\multirow{2}{*}{\begin{tabular}[c]{@{}c@{}}ResNet\\ -20\end{tabular}}} & \multicolumn{1}{c|}{No}  & \multicolumn{1}{c|}{\multirow{2}{*}{13.96}} & \multicolumn{1}{c|}{74.38} & 97.64 & \multicolumn{1}{c|}{99.57 (85.75)} & \multicolumn{1}{c|}{99.44 (84.97)} & 84.38 (16.48) & \multicolumn{1}{c|}{100.0 (83.23)} & \multicolumn{1}{c|}{99.96 (85.86)} & \multicolumn{1}{c|}{100.0 (82.70)} & 100.0 (83.13)                      & \multicolumn{1}{c|}{82.66 (15.74)} & \multicolumn{1}{c|}{99.98 (84.45)} & \multicolumn{1}{c|}{100.0 (87.63)} & 99.73 (73.72)                      & \multicolumn{1}{c|}{100.0 (85.62)} & \multicolumn{1}{c|}{100.0 (83.22)} & \multicolumn{1}{c|}{99.05 (82.67)} & 99.88 (80.53)                      \\ \cline{2-2} \cline{4-20} 
\multicolumn{1}{|c|}{}                                                                      & \multicolumn{1}{c|}{Yes} & \multicolumn{1}{c|}{}                       & \multicolumn{1}{c|}{55.61} & 87.73 & \multicolumn{1}{c|}{77.74 (84.34)} & \multicolumn{1}{c|}{81.80 (82.06)} & 88.09 (76.32) & \multicolumn{1}{c|}{96.03 (57.51)} & \multicolumn{1}{c|}{88.46 (61.83)} & \multicolumn{1}{c|}{94.31 (50.03)} & 94.50 (50.95)                      & \multicolumn{1}{c|}{85.51 (41.56)} & \multicolumn{1}{c|}{89.11 (53.56)} & \multicolumn{1}{c|}{93.78 (47.58)} & 69.94 (20.31)                      & \multicolumn{1}{c|}{99.95 (78.39)} & \multicolumn{1}{c|}{90.30 (36.82)} & \multicolumn{1}{c|}{83.92 (60.79)} & 82.46 (32.16)                      \\ \hline
\end{tabular}
}
}
\label{tab:corr_robust}
\vspace{-1em}
\end{table*}


\subsection{Empirical Analysis of the Posterior Distribution}
{We have performed an empirical evaluation of our method on the CIFAR-10 dataset and the ResNet-18 topology. We have observed that as the number of Monte Carlo samples increases, the histogram of the posterior distribution for each of the 10 classes indeed approaches a Gaussian distribution and can be considered as an approximate Gaussian distribution similar to the MC-Dropout method.}

\begin{figure}
    \centering
\includegraphics[width=1\linewidth]{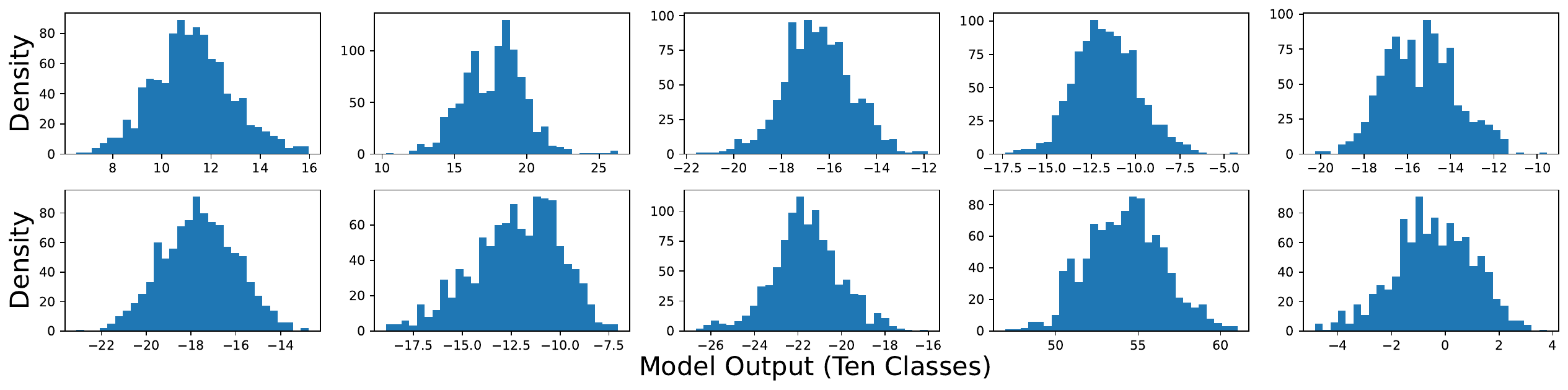}
    \caption{Per class posterior distribution of ResNet-18 topology with a Monte Carlo sample size of 1000.}
    \label{fig:post_dist}
\end{figure}

\section{Conclusion} \label{sec:Conclusion}
In this paper, we propose a novel Dropout approach, Scale-Dropout, which drops the entire scale vector of a layer stochastically with a probability $p$. Our approach required only one Dropout module in the hardware, regardless of the model size. Additionally, we propose scale-Dropout-based Bayesian inference, MC-Scale Dropout, for efficient uncertainty estimation. Furthermore, we propose a novel SOT-MRAM-based CIM implementation for accelerating Bayesian NNs. In our CIM architecture, the stochastic and deterministic aspects of SOT-MRAM have been combined in a crossbar array-based architecture, with only changes made in peripheral circuitry, and achieve up to $100\times$ of energy savings. In terms of uncertainty estimation, our approach can detect up to $100\%$ out-of-distribution data, significantly higher uncertainty estimates compared to popular uncertainty estimation approaches. Additionally, predictive performance is improved by up to $1.33$\% compared to SOTA binary BayNN and improved by up to $0.26$\% compared to conventional BNN approaches. Our approach combines the algorithmic approach with the cost-effective and energy-efficient SOT-MRAM-based CIM implementation for reliable prediction.

\bibliographystyle{IEEEtran}
\typeout{}
\bibliography{references}

\begin{thebibliography}{10}
\providecommand{\url}[1]{#1}
\csname url@samestyle\endcsname
\providecommand{\newblock}{\relax}
\providecommand{\bibinfo}[2]{#2}
\providecommand{\BIBentrySTDinterwordspacing}{\spaceskip=0pt\relax}
\providecommand{\BIBentryALTinterwordstretchfactor}{4}
\providecommand{\BIBentryALTinterwordspacing}{\spaceskip=\fontdimen2\font plus
\BIBentryALTinterwordstretchfactor\fontdimen3\font minus
  \fontdimen4\font\relax}
\providecommand{\BIBforeignlanguage}[2]{{%
\expandafter\ifx\csname l@#1\endcsname\relax
\typeout{** WARNING: IEEEtran.bst: No hyphenation pattern has been}%
\typeout{** loaded for the language `#1'. Using the pattern for}%
\typeout{** the default language instead.}%
\else
\language=\csname l@#1\endcsname
\fi
#2}}
\providecommand{\BIBdecl}{\relax}
\BIBdecl

\bibitem{krizhevsky2012imagenet}
A.~Krizhevsky \emph{et~al.}, ``Imagenet classification with deep convolutional
  neural networks,'' \emph{Advances in neural information processing systems},
  vol.~25, 2012.

\bibitem{devlin2018bert}
J.~Devlin \emph{et~al.}, ``Bert: Pre-training of deep bidirectional
  transformers for language understanding,'' \emph{arXiv preprint
  arXiv:1810.04805}, 2018.

\bibitem{esteva2021deep}
A.~Esteva \emph{et~al.}, ``Deep learning-enabled medical computer vision,''
  \emph{NPJ digital medicine}, vol.~4, p.~5, 2021.

\bibitem{tang2020anomaly}
T.-W. Tang \emph{et~al.}, ``Anomaly detection neural network with dual
  auto-encoders gan and its industrial inspection applications,''
  \emph{Sensors}, vol.~20, p. 3336, 2020.

\bibitem{kendall2017uncertainties}
A.~Kendall and Y.~Gal, ``What uncertainties do we need in bayesian deep
  learning for computer vision?'' \emph{NeurIPS}, vol.~30, 2017.

\bibitem{neal2012bayesian}
R.~M. Neal, \emph{Bayesian learning for neural networks}.\hskip 1em plus 0.5em
  minus 0.4em\relax Springer Science \& Business Media, 2012, vol. 118.

\bibitem{hubara2016binarized}
I.~Hubara \emph{et~al.}, ``Binarized neural networks,'' \emph{NeurIPS},
  vol.~29, 2016.

\bibitem{lakshminarayanan2017simple}
B.~Lakshminarayanan \emph{et~al.}, ``Simple and scalable predictive uncertainty
  estimation using deep ensembles,'' \emph{NeurIPS}, 2017.

\bibitem{blundell2015weight}
C.~Blundell \emph{et~al.}, ``Weight uncertainty in neural network,'' in
  \emph{ICML}.\hskip 1em plus 0.5em minus 0.4em\relax PMLR, 2015.

\bibitem{gal2016Dropout}
Y.~Gal and Z.~Ghahramani, ``Dropout as a bayesian approximation: Representing
  model uncertainty in deep learning,'' in \emph{international conference on
  machine learning}.\hskip 1em plus 0.5em minus 0.4em\relax PMLR, 2016, pp.
  1050--1059.

\bibitem{merolla2014million}
P.~A. Merolla \emph{et~al.}, ``A million spiking-neuron integrated circuit with
  a scalable communication network and interface,'' \emph{Science}, vol. 345,
  pp. 668--673, 2014.

\bibitem{zou2021breaking}
X.~Zou \emph{et~al.}, ``Breaking the von neumann bottleneck: architecture-level
  processing-in-memory technology,'' \emph{Science China Information Sciences},
  vol.~64, p. 160404, 2021.

\bibitem{yu2018neuro}
S.~Yu, ``Neuro-inspired computing with emerging nonvolatile memorys,''
  \emph{Proceedings of the IEEE}, vol. 106, pp. 260--285, 2018.

\bibitem{ahmed_spinbayes_2023}
\BIBentryALTinterwordspacing
S.~T. Ahmed \emph{et~al.}, ``{SpinBayes}: {Algorithm}-{Hardware} {Co}-{Design}
  for {Uncertainty} {Estimation} {Using} {Bayesian} {In}-{Memory}
  {Approximation} on {Spintronic}-{Based} {Architectures},'' \emph{ACM
  Transactions on Embedded Computing Systems}, vol.~22, pp. 131:1--131:25, Sep.
  2023. [Online]. Available: \url{https://doi.org/10.1145/3609116}
\BIBentrySTDinterwordspacing

\bibitem{ahmed2023spindrop}
------, ``Spindrop: Dropout-based bayesian binary neural networks with
  spintronic implementation,'' \emph{IEEE Journal on Emerging and Selected
  Topics in Circuits and Systems}, vol.~13, pp. 150--164, 2023.

\bibitem{soyed_TNANO23}
------, ``Spatial-spindrop: Spatial dropout-based binary bayesian neural
  network with spintronics implementation,'' \emph{arXiv preprint
  arXiv:2306.10185}, 2023.

\bibitem{soyed_nanoarch22}
------, ``Binary bayesian neural networks for efficient uncertainty estimation
  leveraging inherent stochasticity of spintronic devices,'' in
  \emph{NANOARCH'22: 17th ACM International Symposium on Nanoscale
  Architectures}.\hskip 1em plus 0.5em minus 0.4em\relax ACM, 2022, pp. 1--6.

\bibitem{bonnet2023bringing}
D.~Bonnet \emph{et~al.}, ``Bringing uncertainty quantification to the
  extreme-edge with memristor-based bayesian neural networks,'' 2023.

\bibitem{ahmed2023scalable}
S.~T. Ahmed \emph{et~al.}, ``Scalable spintronics-based bayesian neural network
  for uncertainty estimation,'' in \emph{2023 Design, Automation \& Test in
  Europe Conference \& Exhibition (DATE)}.\hskip 1em plus 0.5em minus
  0.4em\relax IEEE, 2023, pp. 1--6.

\bibitem{Dorrance2022}
R.~Dorrance \emph{et~al.}, ``An energy-efficient bayesian neural network
  accelerator with cim and a time-interleaved hadamard digital grng using 22-nm
  finfet,'' \emph{IEEE Journal of Solid-State Circuits}, vol.~58, pp.
  2826--2838, 2023.

\bibitem{rastegari2016xnor}
M.~Rastegari \emph{et~al.}, ``Xnor-net: Imagenet classification using binary
  convolutional neural networks,'' in \emph{European conference on computer
  vision}.\hskip 1em plus 0.5em minus 0.4em\relax Springer, 2016, pp. 525--542.

\bibitem{bulat2019xnor}
A.~Bulat and G.~Tzimiropoulos, ``Xnor-net++: Improved binary neural networks,''
  \emph{arXiv preprint arXiv:1909.13863}, 2019.

\bibitem{qin2020forward}
H.~Qin \emph{et~al.}, ``Forward and backward information retention for accurate
  binary neural networks,'' in \emph{Proceedings of the IEEE/CVF conference on
  computer vision and pattern recognition}, 2020, pp. 2250--2259.

\bibitem{nguyen2015deep}
A.~Nguyen \emph{et~al.}, ``Deep neural networks are easily fooled: High
  confidence predictions for unrecognizable images,'' in \emph{Proceedings of
  the IEEE conference on computer vision and pattern recognition}, 2015, pp.
  427--436.

\bibitem{he2023survey}
W.~He and Z.~Jiang, ``A survey on uncertainty quantification methods for deep
  neural networks: An uncertainty source perspective,'' \emph{arXiv preprint
  arXiv:2302.13425}, 2023.

\bibitem{srivastava2014Dropout}
N.~Srivastava \emph{et~al.}, ``Dropout: a simple way to prevent neural networks
  from overfitting,'' \emph{The journal of machine learning research}, vol.~15,
  pp. 1929--1958, 2014.

\bibitem{wan2013regularization}
L.~Wan \emph{et~al.}, ``Regularization of neural networks using dropconnect,''
  in \emph{International conference on machine learning}.\hskip 1em plus 0.5em
  minus 0.4em\relax PMLR, 2013, pp. 1058--1066.

\bibitem{tompson2015efficient}
J.~Tompson \emph{et~al.}, ``Efficient object localization using convolutional
  networks,'' in \emph{Proceedings of the IEEE conference on computer vision
  and pattern recognition}, 2015, pp. 648--656.

\bibitem{kingma2015variational}
D.~P. Kingma \emph{et~al.}, ``Variational dropout and the local
  reparameterization trick,'' \emph{Advances in neural information processing
  systems}, vol.~28, 2015.

\bibitem{bishop2006pattern}
C.~M. Bishop and N.~M. Nasrabadi, \emph{Pattern recognition and machine
  learning}.\hskip 1em plus 0.5em minus 0.4em\relax Springer, 2006, vol.~4,
  no.~4.

\bibitem{dieny_opportunities_2020}
B.~Dieny \emph{et~al.}, ``Opportunities and challenges for spintronics in the
  microelectronics industry,'' \emph{Nature Electronics}, vol.~3, Aug. 2020.

\bibitem{lee_world-most_2022}
T.~Y. Lee \emph{et~al.}, ``World-most energy-efficient {MRAM} technology for
  non-volatile {RAM} applications,'' in \emph{2022 {International} {Electron}
  {Devices} {Meeting} ({IEDM})}.\hskip 1em plus 0.5em minus 0.4em\relax IEEE,
  Dec. 2022, pp. 10.7.1--10.7.4, iSSN: 2156-017X.

\bibitem{shao_roadmap_2021}
Q.~Shao \emph{et~al.}, ``Roadmap of spin-orbit torques,'' \emph{IEEE TransMag},
  2021.

\bibitem{awano_bynqnet_2020}
H.~Awano and M.~Hashimoto, ``Bynqnet: Bayesian neural network with quadratic
  activations for sampling-free uncertainty estimation on fpga,'' in \emph{2020
  Design, Automation \& Test in Europe Conference \& Exhibition (DATE)}.\hskip
  1em plus 0.5em minus 0.4em\relax IEEE, 2020, pp. 1402--1407.

\bibitem{fan_high-performance_2021}
H.~Fan \emph{et~al.}, ``High-{Performance} {FPGA}-based {Accelerator} for
  {Bayesian} {Neural} {Networks},'' in \emph{2021 58th {ACM}/{IEEE} {Design}
  {Automation} {Conference} ({DAC})}.\hskip 1em plus 0.5em minus 0.4em\relax
  San Francisco, CA, USA: IEEE Press, Dec. 2021, pp. 1063--1068.

\bibitem{fan_fpga-based_2022}
------, ``{FPGA}-{Based} {Acceleration} for {Bayesian} {Convolutional} {Neural}
  {Networks},'' \emph{IEEE Transactions on Computer-Aided Design of Integrated
  Circuits and Systems}, vol.~41, pp. 5343--5356, Dec. 2022, conference Name:
  IEEE Transactions on Computer-Aided Design of Integrated Circuits and
  Systems.

\bibitem{malhotra_exploiting_2020}
A.~Malhotra \emph{et~al.}, ``Exploiting {Oxide} {Based} {Resistive} {RAM}
  {Variability} for {Bayesian} {Neural} {Network} {Hardware} {Design},''
  \emph{IEEE Transactions on Nanotechnology}, vol.~19, pp. 328--331, 2020,
  conference Name: IEEE Transactions on Nanotechnology.

\bibitem{dalgaty_situ_2021}
T.~Dalgaty \emph{et~al.}, ``\BIBforeignlanguage{en}{In situ learning using
  intrinsic memristor variability via {Markov} chain {Monte} {Carlo}
  sampling},'' \emph{\BIBforeignlanguage{en}{Nature Electronics}}, vol.~4, pp.
  151--161, Feb. 2021, number: 2 Publisher: Nature Publishing Group.

\bibitem{yang_all-spin_2020}
K.~Yang \emph{et~al.}, ``All-spin bayesian neural networks,'' \emph{IEEE
  Transactions on Electron Devices}, vol.~67, pp. 1340--1347, 2020.

\bibitem{lu_algorithm-hardware_2022}
A.~Lu \emph{et~al.}, ``An {Algorithm}-{Hardware} {Co}-{Design} for {Bayesian}
  {Neural} {Network} {Utilizing} {SOT}-{MRAM}’s {Inherent} {Stochasticity},''
  \emph{IEEE-JXCDC}, 2022.

\bibitem{ioffe2015batch}
S.~Ioffe and C.~Szegedy, ``Batch normalization: Accelerating deep network
  training by reducing internal covariate shift,'' in \emph{International
  conference on machine learning}.\hskip 1em plus 0.5em minus 0.4em\relax pmlr,
  2015, pp. 448--456.

\bibitem{lee_emerging_2016}
S.-W. Lee \emph{et~al.}, ``Emerging {Three}-{Terminal} {Magnetic} {Memory}
  {Devices},'' \emph{Proceedings of the IEEE}, vol. 104, pp. 1831--1843, Oct.
  2016, conference Name: Proceedings of the IEEE.

\bibitem{gokmen_training_2017}
T.~Gokmen \emph{et~al.}, ``Training deep convolutional neural networks with
  resistive cross-point devices,'' \emph{Frontiers in neuroscience}, vol.~11,
  p. 538, 2017.

\bibitem{peng_optimizing_2019}
X.~Peng \emph{et~al.}, ``Optimizing weight mapping and data flow for
  convolutional neural networks on rram based processing-in-memory
  architecture,'' in \emph{IEEE ISCAS}.\hskip 1em plus 0.5em minus 0.4em\relax
  IEEE, 2019, pp. 1--5.

\bibitem{al2020dataset}
W.~Al-Dhabyani \emph{et~al.}, ``Dataset of breast ultrasound images,''
  \emph{Data in brief}, vol.~28, p. 104863, 2020.

\bibitem{ma2020covid}
J.~Ma \emph{et~al.}, ``Covid-19 ct lung and infection segmentation dataset.
  zenodo,'' 2020.

\bibitem{mendoncca2013ph}
T.~Mendon{\c{c}}a \emph{et~al.}, ``Ph 2-a dermoscopic image database for
  research and benchmarking.''\hskip 1em plus 0.5em minus 0.4em\relax IEEE,
  2013, pp. 5437--5440.

\bibitem{netzer2011reading}
Y.~Netzer \emph{et~al.}, ``Reading digits in natural images with unsupervised
  feature learning,'' 2011.

\bibitem{coates2011analysis}
A.~Coates \emph{et~al.}, ``An analysis of single-layer networks in unsupervised
  feature learning,'' in \emph{Proceedings of the fourteenth international
  conference on artificial intelligence and statistics}.\hskip 1em plus 0.5em
  minus 0.4em\relax JMLR Workshop and Conference Proceedings, 2011, pp.
  215--223.

\bibitem{he2016deep}
K.~He \emph{et~al.}, ``Deep residual learning for image recognition,'' in
  \emph{Proceedings of the IEEE conference on computer vision and pattern
  recognition}, 2016, pp. 770--778.

\bibitem{simonyan2014very}
K.~Simonyan and A.~Zisserman, ``Very deep convolutional networks for
  large-scale image recognition,'' \emph{arXiv preprint arXiv:1409.1556}, 2014.

\bibitem{ronneberger2015u}
O.~Ronneberger \emph{et~al.}, ``U-net: Convolutional networks for biomedical
  image segmentation,'' in \emph{International Conference on Medical image
  computing and computer-assisted intervention}.\hskip 1em plus 0.5em minus
  0.4em\relax Springer, 2015, pp. 234--241.

\bibitem{BMVC2017_57}
V.~B. Alex~Kendall and R.~Cipolla, ``Bayesian segnet: Model uncertainty in deep
  convolutional encoder-decoder architectures for scene understanding,'' in
  \emph{Proceedings of the British Machine Vision Conference (BMVC)}, 2017, pp.
  57.1--57.12.

\bibitem{choi2018pact}
J.~Choi \emph{et~al.}, ``Pact: Parameterized clipping activation for quantized
  neural networks,'' \emph{arXiv preprint arXiv:1805.06085}, 2018.

\bibitem{danouchi_spin_2022}
K.~Danouchi \emph{et~al.}, ``Spin {Orbit} {Torque}-based {Crossbar} {Array} for
  {Error} {Resilient} {Binary} {Convolutional} {Neural} {Network},'' in
  \emph{{23RD} {IEEE} {LATIN}-{AMERICAN} {TEST} {SYMPOSIUM}}, Montevideo,
  Uruguay, Sep. 2022.

\bibitem{nvsim6218223}
X.~Dong \emph{et~al.}, ``Nvsim: A circuit-level performance, energy, and area
  model for emerging nonvolatile memory,'' \emph{IEEE Transactions on
  Computer-Aided Design of Integrated Circuits and Systems}, vol.~31, pp.
  994--1007, 2012.

\bibitem{rad}
R.~Ding \emph{et~al.}, ``Regularizing activation distribution for training
  binarized deep networks,'' in \emph{Proceedings of the IEEE/CVF conference on
  computer vision and pattern recognition}, 2019, pp. 11\,408--11\,417.

\bibitem{dorefa}
S.~Zhou \emph{et~al.}, ``Dorefa-net: Training low bitwidth convolutional neural
  networks with low bitwidth gradients,'' \emph{arXiv preprint
  arXiv:1606.06160}, 2016.

\bibitem{dsq}
R.~Gong \emph{et~al.}, ``Differentiable soft quantization: Bridging
  full-precision and low-bit neural networks,'' in \emph{Proceedings of the
  ICCV}, 2019, pp. 4852--4861.

\bibitem{LAB}
L.~Hou \emph{et~al.}, ``Loss-aware binarization of deep networks,'' in
  \emph{International Conference on Learning Representations (ICLR)}, 2017.

\bibitem{mobiny2021dropconnect}
A.~Mobiny \emph{et~al.}, ``Dropconnect is effective in modeling uncertainty of
  bayesian deep networks,'' \emph{Scientific reports}, 2021.

\bibitem{cai_vibnn_2018}
R.~Cai \emph{et~al.}, ``Vibnn: Hardware acceleration of bayesian neural
  networks,'' \emph{ACM SIGPLAN Notices}, vol.~53, pp. 476--488, 2018.

\bibitem{jia_efficient_2021}
X.~Jia \emph{et~al.}, ``Efficient {Computation} {Reduction} in {Bayesian}
  {Neural} {Networks} {Through} {Feature} {Decomposition} and {Memorization},''
  \emph{IEEE Trans. on Neural Networks and Learning Systems}, vol.~32, Apr.
  2021.

\bibitem{hendrycks2019benchmarking}
D.~Hendrycks and T.~Dietterich, ``Benchmarking neural network robustness to
  common corruptions and perturbations,'' \emph{arXiv preprint
  arXiv:1903.12261}, 2019.

\bibitem{ielmini_status_2022}
D.~Ielmini \emph{et~al.}, ``Status and challenges of in-memory computing for
  neural accelerators,'' in \emph{2022 {International} {Symposium} on {VLSI}
  {Technology}, {Systems} and {Applications} ({VLSI}-{TSA})}, Apr. 2022, pp.
  1--2, iSSN: 1930-8868.

\end{thebibliography}

\vskip -2\baselineskip plus -1fil
\begin{IEEEbiography}[
{\includegraphics[width=1in,height=1.25in, clip, keepaspectratio]{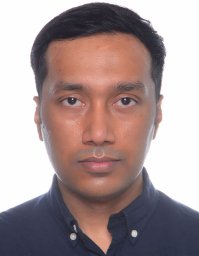}}]{Soyed Tuhin Ahmed} has received his bachelor's in Electrical and Electronics Engineering from American International University Bangladesh with Summa Cum Laude and afterwards Master in Communication Engineering from Technische Universität München in 2020. He joined the CDNC group at Karlsruhe Institute of Technology, Karlsruhe, Germany as a PhD student in September 2020. His current research interests are Deep learning, scalable and low-cost uncertainty estimation, resilient hardware accelerator for machine learning, and robust and accurate deep learning. 
\end{IEEEbiography}
\vskip -2\baselineskip plus -1fil
\begin{IEEEbiography}
[{\includegraphics[width=1in,height=1.25in,clip, keepaspectratio]{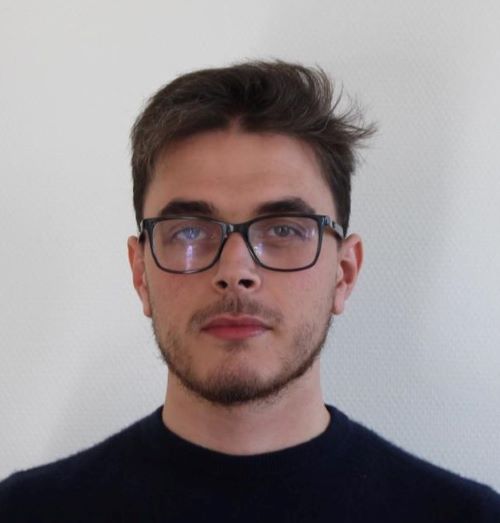}}]{Kamal Danouchi} graduated from the University of Aix-Marseille, France, with an engineering degree in Microelectronics in 2021. He is currently pursuing his PhD at CEA-SPINTEC. His research interests cover emerging non-volatile memories, spintronics, and IC design for unconventional computing. 
\end{IEEEbiography}
\vskip -2\baselineskip plus -1fil

\begin{IEEEbiography}
[{\includegraphics[width=1in,height=1.25in,clip,keepaspectratio]{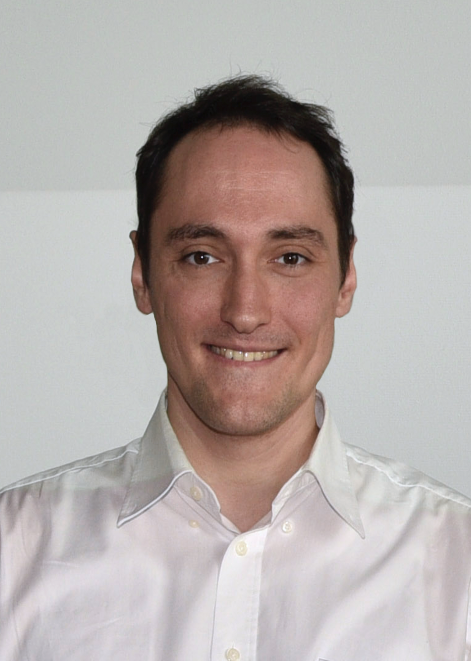}}]{Michael Hefenbrock} received the Ph.D. degree in computer science from the Karlsruhe Institute of Technology (KIT), Karlsruhe, Germany, in 2022. He is currently the head of artificial intelligence at RevoAI GmbH, Karlsruhe. His current research interests include machine learning and optimization and their application to problems in design automation.
\end{IEEEbiography}
\vskip -2\baselineskip plus -1fil

\begin{IEEEbiography}
[{\includegraphics[width=1in,height=1.25in,clip,keepaspectratio]{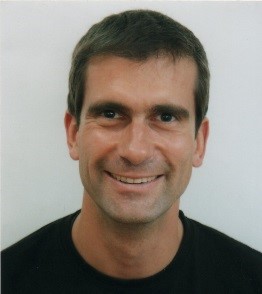}}]{Guillaume Prenat} graduated from Grenoble Institute of Technology in France, where he obtained his engineer degree in 2002 and his PhD degree in microelectronics in 2005. He joined SPINTEC in 2006 to take in charge the design activity.  In 2021, he obtained his habilitation to conduct research from University Grenoble Alpes. His field of interest covers the development of design tools for CMOS/magnetic technology and the evaluation of hybrid non-volatile circuits (FPGA, processors...) to contribute to circumventing the limits of microelectronics. In this framework, he was involved as the scientific contact in 8 European and French research projects, and as the coordinator of a H2020 ICT project embedding 9 academic and industrial partners. He was also in charge of the collaboration contract between Spintec and the startup eVaderis.
\end{IEEEbiography}
\vskip -2\baselineskip plus -1fil

\begin{IEEEbiography}
[{\includegraphics[width=1in,height=1.25in,clip,keepaspectratio]{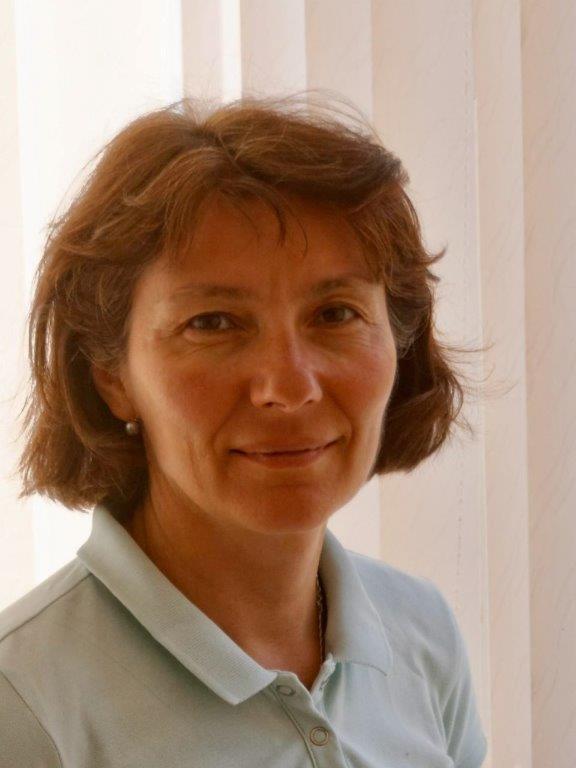}}]{Lorena Anghel}  received the Ph.D. degree (cum laude) from Grenoble INP in 2000. From 2016 to 2020, she was the Vice President of Grenoble INP, in charge of industrial relationships, where she is currently the Scientific Director. She is also a Full Professor at Grenoble INP and a member of the Research Staff at the Spintec Laboratory. She has published more than 130 publications in international conferences and symposia. 
She was a recipient of several best paper and outstanding paper awards. She had fulfilled positions, such as the General Chair and Program Chair for many prestigious IEEE conferences, including IEEE VTS, IEEE ETS, IEEE NANOARCH, and IEEE On-Line Test Symposium.
\end{IEEEbiography}
\vskip -2\baselineskip plus -1fil
\begin{IEEEbiography}
[{\includegraphics[width=1in,height=1.25in,clip,keepaspectratio]{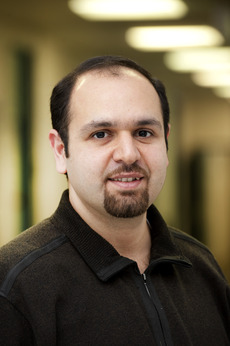}}]{Mehdi B. Tahoori} (M'03, SM'08, F'21) received the B.S. degree in computer engineering from the Sharif University of Technology, Tehran, Iran, in 2000, and the M.S. and Ph.D. degrees in electrical engineering from Stanford University, Stanford, CA, in 2002 and 2003, respectively. He is currently a Full Professor 
at the Karlsruhe Institute of Technology, Karlsruhe, Germany. 
In 2003, he was an Assistant Professor at the Department of Electrical and Computer Engineering, Northeastern University, where he became an Associate Professor in 2009. From August to December 2015, he was a visiting professor at the VLSI Design and Education Center (VDEC), University of Tokyo, Japan. From 2002 to 2003, he was a Research Scientist with Fujitsu Laboratories of America, Sunnyvale, CA. 
Prof. Tahoori was a recipient of the National Science Foundation Early Faculty Development (CAREER) Award. He has received a number of best paper awards at various conferences and journals, including ICCAD, FPL, TODAES, and TVLSI. He is a fellow of the IEEE and a recipient of the European Research Council (ERC) Advanced Grant. 
\end{IEEEbiography}

\end{document}